\pdfoutput=1

\documentclass[11pt]{article}

\usepackage[preprint]{acl}

\usepackage{times}
\usepackage{latexsym}

\usepackage[T1]{fontenc}

\usepackage[utf8]{inputenc}

\usepackage{microtype}

\usepackage{inconsolata}

\usepackage{tabularx} 
\usepackage{listings}
\usepackage{tcolorbox}
\usepackage{times}
\usepackage{soul}
\usepackage{url}

\usepackage{booktabs}
\usepackage{tabularx}
\usepackage{multirow}
\usepackage{array}
\usepackage{placeins}
\usepackage{listings}

\usepackage{graphicx}
\setkeys{Gin}{draft=false} 
\usepackage{amsmath}
\usepackage{amsthm}
\usepackage{booktabs}
\usepackage{algorithm}
\usepackage{algorithmic}
\usepackage{inconsolata}
\usepackage{import}
\usepackage{microtype}
\usepackage{layout}
\usepackage{tabularx, makecell}
\usepackage{booktabs}
\usepackage{mathrsfs}
\usepackage{amssymb} 
\usepackage{url}
\usepackage{graphicx}
\usepackage{bbm}
\usepackage{xspace,paralist}
\usepackage{times,latexsym}
\usepackage{amsmath}
\usepackage{appendix}
\usepackage{comment} 
\usepackage{enumitem}
\usepackage{makecell}
\usepackage{multirow}
\usepackage{xcolor}
\usepackage{fontawesome5}
\usepackage{graphicx}
\usepackage{cleveref}
\usepackage{tcolorbox}
\usepackage{adjustbox}
\usepackage{relsize}
\usepackage{todonotes}
\usepackage{natbib}
\usepackage{CJKutf8}
\usepackage{colortbl}
\usepackage{subcaption}
\usepackage{booktabs}
\usepackage{adjustbox}
\usepackage{paralist}

\definecolor{lightyellow}{RGB}{255,249,230} 
\definecolor{barblue}{HTML}{4A90D9}
\definecolor{rowgray}{gray}{0.95}
\definecolor{best}{HTML}{D6EAF8}       
\definecolor{degradelight}{HTML}{FADBD8}   
\definecolor{degradestrong}{HTML}{F1948A}  
\definecolor{improve}{HTML}{D5F5E3}        
\definecolor{degradestrong}{HTML}{F1948A}  
\definecolor{warnlight}{HTML}{FEF9E7}      
\definecolor{warnstrong}{HTML}{F9E79F}     

\newcommand{\ibar}[1]{%
  \rlap{\textcolor{barblue}{\rule{#1pt}{7pt}}}%
  \hspace{70pt}
}

\usepackage{amssymb}
\usepackage{pifont}
\usepackage{booktabs}
\usepackage{listings}
\usepackage{xcolor}
\usepackage{tcolorbox}
\usepackage[switch]{lineno}
\usepackage{url}

\usepackage{colortbl, xcolor, multirow, booktabs, adjustbox}
\definecolor{yellow}{RGB}{255, 255, 150}      
\definecolor{lightblue}{RGB}{173, 216, 230}   
\definecolor{lightred}{RGB}{255, 182, 193}    
\definecolor{lightgreen}{RGB}{144, 238, 144}  

\definecolor{greenbox}{RGB}{144, 238, 144}    
\definecolor{redbox}{RGB}{255, 182, 193}      
\definecolor{bluebox}{RGB}{135, 206, 235}     
\definecolor{yellowbox}{RGB}{255, 255, 0}     
\definecolor{posgreen}{RGB}{198, 239, 206}  
\definecolor{negred}{RGB}{255, 199, 206}    

\usepackage{multirow}
\usepackage{graphicx}
\usepackage{tcolorbox}
\usepackage{xcolor}
\usepackage{CJKutf8} 
\usepackage[utf8]{inputenc}

\usepackage[ruled,linesnumbered,algo2e]{algorithm2e}

\tcbset{
    eval_prompt/.style={
        colback=gray!5,
        colframe=black!70,
        boxrule=0.5pt,
        arc=2mm,
        left=8pt, right=8pt, top=8pt, bottom=8pt,
        fontupper=\small,
        width=\linewidth
    }
}

\title{Beyond 0--100: How Confidence Scale Design Shapes Verbalized Confidence and LLM Metacognition}

\author{
  Yuyang Dai \\
  INSAIT \\
  \texttt{y9657422@gmail.com}
  \And
  Yuxia Wang \\
  INSAIT \\
  \texttt{yuxia.wang@insait.ai}
  \AND
  \normalfont\small
  \faGlobe\enspace\href{https://yuyang915.github.io/confidence-scale/}{\texttt{yuyang915.github.io/confidence-scale/}}\\[-1pt]
  \faGithub\enspace\href{https://github.com/yuyang915/confidence-scale}{\texttt{github.com/yuyang915/confidence-scale}}
}

\hypersetup{
  pdftitle={Beyond 0--100: How Confidence Scale Design Shapes Verbalized Confidence and LLM Metacognition},
  pdfauthor={Yuyang Dai; Yuxia Wang},
  pdfsubject={Confidence scale design, verbalized confidence, calibration, and metacognition in large language models},
  pdfkeywords={large language models, LLM calibration, verbalized confidence, confidence scale, metacognition, uncertainty estimation, expected calibration error, meta-d-prime},
  pdfinfo={ProjectURL={https://yuyang915.github.io/confidence-scale/},Repository={https://github.com/yuyang915/confidence-scale}}
}

\begin{document}
\maketitle
\begin{abstract}
Verbalized confidence, in which LLMs report a numerical certainty score, is widely used for uncertainty estimation and calibration in black-box settings, yet the confidence scale itself (typically 0--100) is rarely examined.
We show that this design choice is not neutral. 
Across six primary LLMs and three datasets, with an additional dense-architecture model used for robustness analysis, we find that verbalized confidence is heavily discretized: the three most frequent values account for 78.2\%--92.1\% of responses under the standard scale. 
To investigate this phenomenon, we manipulate confidence scales along three dimensions: granularity, boundary placement, and range regularity, and evaluate metacognitive sensitivity using $meta\text{-}d'$. 
Within the evaluated models, datasets, prompts, and decoding conditions, a 0--20 scale improves metacognitive efficiency over the standard 0--100 format, while aggressive boundary compression degrades performance and round-number preferences persist even under irregular ranges. 
These results show that confidence scale design affects the quality of verbalized uncertainty in our experiments and should be treated as a first-class experimental variable in LLM evaluation.
\end{abstract}

\section{Introduction}
As Large Language Models (LLMs) are increasingly integrated into decision-making pipelines, reliably estimating their confidence or uncertainty becomes critical. In black-box settings, \textit{\textbf{verbalized confidence}}, where models report a numerical certainty score via prompting, has become the predominant paradigm \citep{xiong2024can}: confidence scores are used to decide when to abstain from answering in selective prediction \citep{kadavath2022language}, to trigger self-correction on low-confidence outputs \citep{li2024confidence}, and to calibrate how much end users should trust a given response \citep{zhou2024relying}. Yet the reliability of these confidence signals remains poorly understood.

Given its ubiquity, research on verbalized confidence has focused on elicitation strategies such as Chain-of-Thought prompting \citep{wei2022chain}, self-consistency \citep{wang2022self}, and training models to express uncertainty \citep{lin2022teaching}, while treating the \textit{\textbf{confidence scale}} itself (typically 0--100) as a neutral instrument. Yet psychometric research has long established that scale granularity and anchoring alter the quality of self-assessment \citep{lozano2008effect, preston2000optimal}. Whether analogous effects hold for LLM judges remains unexplored.

We find that LLM confidence reports exhibit a striking phenomenon we term \textit{confidence discretization}: LLMs do not use the 0--100 scale as a continuous spectrum but cluster outputs around a small set of round-number anchors. 
This compression, which we find to be pervasive yet model-specific, suggests that verbalized confidence may be shaped as much by token-level biases \citep{zhao2021calibrate, shaki2023cognitive} as by genuine self-evaluation, potentially distorting standard calibration metrics such as ECE \citep{guo2017calibration}. This raises a central question: \textit{does confidence scale design affect the quality of verbalized uncertainty signals in LLMs?}

To investigate, we present the first systematic empirical study of confidence scale design for LLM metacognition, manipulating scales along three dimensions: \textit{granularity} (0--5 to 0--100), \textit{boundary shifting} (e.g.,\ lower bound raised to 20 or 60 \citep{tversky1974judgment}), and \textit{semantic robustness} via non-standard ranges (e.g.,\ 0--73, 14--86 \citep{pellert2024ai}). We adopt $meta\text{-}d'$ from signal detection theory \citep{maniscalco2012signal, fleming2014measure} to quantify metacognitive sensitivity independent of overall bias.
Our contributions are:
\begin{itemize}
    \item We reveal the vulnerability of verbalized confidence to scale design, identifying \textbf{confidence discretization}, a robust phenomenon across multiple LLM families.
    \item Confidence scale granularity significantly modulates metacognitive quality, identifying a consistent sweet spot 0--20 that outperforms the standard 0--100 scale.
    \item Through non-standard scales, we show that LLMs exhibit limited adaptation to numerical range boundaries (\textit{e.g.,\ raising the lower bound to 60 as $[60,100]$ leaves models clustered near the ceiling}), providing practical guidelines for confidence scale design.
\end{itemize}

\begin{figure*}[t!]
\centering
\includegraphics[width=0.95\linewidth]{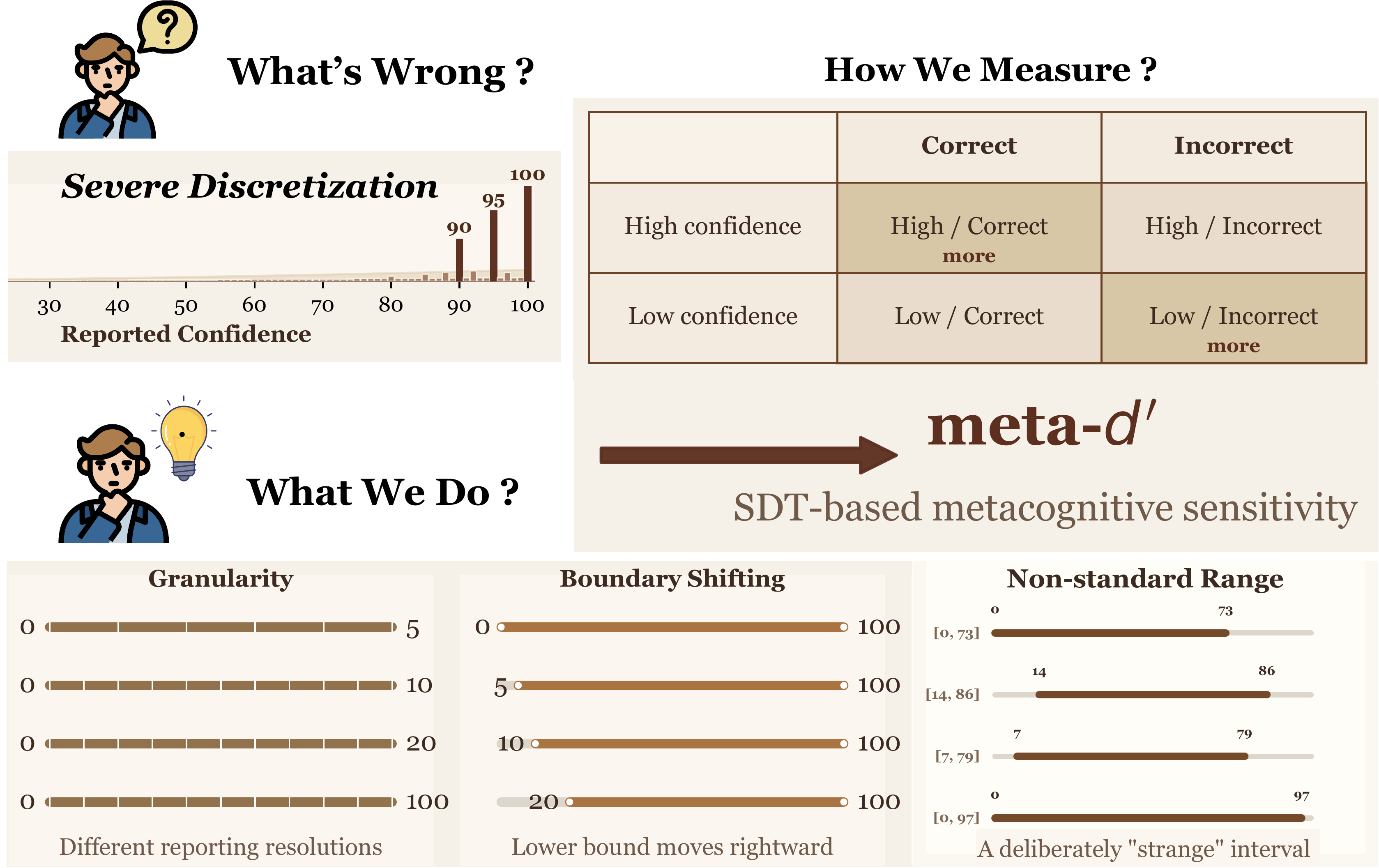}
\caption{Confidence discretization phenomenon, $meta\text{-}d'$ evaluation, and three-dimensional scale manipulations.}
\label{fig:overview}
\end{figure*}

\section{Related Work}

\paragraph{Verbalized Confidence and Calibration}
As LLMs became accessible primarily through black-box APIs, uncertainty estimation shifted from logit-based methods \citep{kadavath2022language} to verbalized confidence, where models report certainty as a numerical score \citep{lin2022teaching, xiong2024can}. Prior work has focused on improving elicitation, via Chain-of-Thought prompting \citep{wei2022chain}, metacognitive prompting \citep{wang2024metacognitive}, self-consistency \citep{wang2022self}, and RLHF-specific strategies \citep{tian2023just}, and on evaluation via Expected Calibration Error \citep[ECE:][]{guo2017calibration}. This literature treats the confidence scale itself as a fixed, neutral backdrop.
However, we find that LLMs do not use the 0--100 scale as a continuous spectrum but cluster outputs around a small set of round-number anchors. 
Then, when models concentrate the vast majority of their reports on two or three values, ECE bins become sparse or empty across most of the range, making calibration estimates unreliable. This \textit{discretization bias} and its dependence on scale design have not been systematically studied.

\paragraph{Scale Design in Psychometrics}
Scale design is known to be non-neutral for human respondents: the number of response categories affects reliability and validity \citep{lozano2008effect, preston2000optimal}, and boundary placement biases judgments toward specific anchors \citep{tversky1974judgment, harvey1997confidence}. Whether such effects transfer to LLMs is unclear, since their outputs are shaped by token-level distributions, potentially reflecting an additional mechanism absent in humans: the statistical prevalence of numerical tokens in pretraining corpora \citep{zhao2021calibrate, shaki2023cognitive}.

\paragraph{Metacognitive Sensitivity and Signal Detection Theory}
Evaluating metacognition requires distinguishing Type-1 performance (task accuracy) from Type-2 sensitivity (the ability to monitor that accuracy) \citep{galvin2003type}. Signal Detection Theory (SDT) provides a principled framework for this decomposition \citep{hautus2021detection}. The $meta\text{-}d'$ metric \citep{maniscalco2012signal} quantifies metacognitive sensitivity by measuring how well confidence ratings distinguish correct from incorrect answers, independently of response bias \citep{fleming2014measure, rausch2023measures}. Standard SDT corrections for extreme proportions ensure robust estimation under skewed response distributions~\citep{hautus1995corrections}. Though recent NLP work has begun disentangling metacognitive ability from task performance \citep{wang2025decoupling}, no study has used SDT-based metrics to examine how confidence scale design influences LLM metacognition. By applying $meta\text{-}d'$ across controlled confidence scales, we address this gap.
\section{Methodology}
\label{sec:methodology}


We frame confidence elicitation as a Type-2 (metacognitive) judgment layered on top of a Type-1 (task-level) decision. Given a task instance $x$ from a dataset $\mathcal{D}$, a model $M$ generates an answer $\hat{y}$ and a verbalized confidence score $c$. The confidence $c$ is constrained within a prompted scale $\mathcal{S} = [l, u]$, where $l$ and $u$ denote the lower and upper bounds respectively, and the model is instructed to report an integer $c \in \{l, l{+}1, \ldots, u\}$. We let $a \in \{0, 1\}$ represent the correctness of $\hat{y}$ against the ground truth $y$.
To enable fair comparison across scales of different ranges, all confidence scores are linearly normalized to the unit interval:
\begin{equation*}
    \hat{c} = \frac{c - l}{u - l}
\end{equation*}
All subsequent metrics are computed over $\hat{c} \in [0, 1]$. Our central research question is: \textit{how does the metacognitive informativeness of $\hat{c}$, defined as its ability to discriminate between $a{=}1$ and $a{=}0$, vary with the \textbf{confidence scale} design $\mathcal{S}$?}

\subsection{Scale Design Dimensions}


\paragraph{Granularity ($\mathcal{G}$)}
We vary the granularity of the confidence scale across five levels: $\mathcal{S} \in \{[0,5],\;[0,10],\;[0,20],\;[0,50],\;[0,100]\}$. This dimension tests a fundamental trade-off. Finer-grained scales offer higher theoretical resolution to express uncertainty, but may also amplify token-level biases by providing more opportunities to default to round-number anchors. Conversely, coarser scales force the model to commit to broader confidence bins, potentially reducing noise at the cost of resolution.

\paragraph{Boundary Shifting ($\mathcal{B}$)}
We fix the upper bound $u{=}100$ and progressively raise the lower bound: $\mathcal{S} \in \{[0,100],\;[20,100],\;[40,100],\;[60,100]\}$. This probes anchoring effects: when the available range is compressed from above, does the model redistribute its confidence to utilize the full new range, or does it remain clustered near the ceiling~\citep{tversky1974judgment}? A failure to redistribute would indicate that the model's high-confidence outputs are driven by token preference.

\paragraph{Non-standard Ranges ($\mathcal{N}$)}
Our range selection follows four principles: \emph{(i)} all bounds avoid conventional multiples of 5 and 10, to eliminate effects from pretraining and encourage reasoning based on scale semantics; 
\emph{(ii)} both lower and upper bounds vary to avoid relying on a single boundary configuration; 
\emph{(iii)} include ranges that contain prominent internal round anchors (e.g.,\ 50, 70), while others largely do not, allowing us to test whether models gravitate toward these anchors independently of the scale boundaries; and \emph{(iv)} include a near-standard range to isolate the effect of boundary irregularity from range width.

Concretely, we organize these conditions into two groups. The first holds the range width approximately constant ($\approx$\,70--73 units) while varying both bounds, isolating the effect of boundary placement:
%
$[0,\,73]$, $[14,\,86]$, $[7,\,79]$. 
%
The second group varies the range width to test the robustness of our findings and introduces diagnostic contrasts:
%
$[3,\,38]$ is narrow with few internal round anchors; $[0,\,97]$ is a near-standard range (minimal pair with $[0,\,100]$).
%
If a model instructed to report confidence in $[0,\,73]$ still clusters at values such as 50 or 70, it indicates reliance on pre-trained numerical heuristics; if reports spread across the admissible range (e.g.,\ 55, 68, 73), it suggests genuine semantic adaptation to the scale constraints. The narrow range $[3,\,38]$, which contains few salient round-number anchors, provides the strongest diagnostic test of this distinction.

\subsection{Evaluation Metrics}


\paragraph{Expected Calibration Error (ECE)}
Following \citet{guo2017calibration}, we partition the normalized confidence space $[0, 1]$ into $B$ equal-width bins and compute:
\begin{equation*}
    \text{ECE} = \sum_{b=1}^{B} \frac{n_b}{N} \left| \text{acc}(b) - \text{conf}(b) \right|
\end{equation*}
where $n_b$ is the number of samples in bin $b$, $N$ is the total sample count, and $\text{acc}(b)$ and $\text{conf}(b)$ are the average accuracy and confidence within that bin. We set $B{=}10$ as the default and additionally report results with $B{=}15$ and adaptive binning in Appendix~\ref{app:ece_bins}, since the choice of $B$ is particularly consequential when confidence distributions are heavily discretized.

\paragraph{AUROC}
We report the Area Under the Receiver Operating Characteristic curve, which measures the model's ability to assign higher confidence to correct predictions than to incorrect ones (failure prediction). Unlike ECE, AUROC is threshold-free and thus less sensitive to discretization artifacts.

\paragraph{Metacognitive Sensitivity ($meta\text{-}d'$)}
To isolate the quality of the metacognitive signal from overall response bias, we adopt $meta\text{-}d'$ from Signal Detection Theory \citep{maniscalco2012signal, fleming2014measure}. The intuition is as follows: in the Type-1 task, a model's ability to distinguish correct from incorrect answers is captured by $d' = \Phi^{-1}(H_1) - \Phi^{-1}(F_1)$, where $H_1 = P(\hat{c} > t \mid a = 1)$ and $F_1 = P(\hat{c} > t \mid a = 0)$ are the Type-1 hit and false-alarm rates, computed by binarizing $\hat{c}$ at the model's own Type-1 decision criterion. Type-2 (metacognitive) task measures \textit{how well does the model's confidence separate its own correct and incorrect responses?} To formalize this, we binarize $\hat{c}$ at a threshold $t$ to obtain ``high confidence'' and ``low confidence'' judgments, then compute:
\begin{compactitem}
    \item \textit{Type-2 Hit Rate}: $H_2 = P(\hat{c} > t \mid a = 1)$, the proportion of correct responses assigned high confidence.
    \item \textit{Type-2 False Alarm Rate}: $F_2 = P(\hat{c} > t \mid a = 0)$, the proportion of incorrect responses assigned high confidence.
\end{compactitem}
Formally, $meta\text{-}d'$ is the value satisfying $\Phi^{-1}(H_2) - \Phi^{-1}(F_2) = meta\text{-}d'$, where $\Phi^{-1}$ denotes the inverse of the standard normal cumulative distribution function, which converts hit and false-alarm \textit{rates} into $z$-scores under the equal-variance Gaussian assumption of SDT,
where the Type-2 criteria are derived by mapping the Type-1 criterion proportionally onto the $meta\text{-}d'$ scale \citep{maniscalco2012signal}. A model with perfect metacognition would have $meta\text{-}d' = d'$; a model whose confidence carries no information about correctness would have $meta\text{-}d' = 0$. We further report the metacognitive efficiency ratio $M_{ratio} = meta\text{-}d' / d'$ \citep{fleming2014measure}, which normalizes for differences in task difficulty across datasets: $M_{ratio} = 1$ indicates that the metacognitive system captures all available Type-1 information, while $M_{ratio} < 1$ indicates metacognitive loss. We use two alternative thresholds ($t = 0.9$ and $t = 0.95$ after normalization) and report results for both. To handle the prevalence of extreme confidence values (e.g.,\ 100\%), we apply the log-linear correction of adding $0.5$ to all cells of the Type-2 contingency table \citep{hautus1995corrections}.

\paragraph{Distribution Diagnostics}
To quantify discretization artifacts under different scale designs, we introduce two additional diagnostic metrics.

\textbf{\textit{Round}} measures the proportion of confidence reports that fall on multiples of five (e.g.,\ 0, 5, 10, ..., 100), capturing models' tendency to rely on round-number anchors.

\textbf{\textit{Viol.}} measures the proportion of responses whose reported confidence lies outside the valid range $[l,u]$ by more than 5\% of the scale width.

\subsection{Experimental Setup}

\paragraph{Models}
We evaluate six LLMs spanning different training paradigms, model families, and scales.
\begin{itemize}
    \item \textbf{Closed-source:} GPT-5.2 \citep{openai2025gpt5} and Gemini 3.1 Pro \citep{google2026gemini3}, represent two frontier commercial APIs.
    \item \textbf{Open-weights (Meta):} LLaMA-4-Maverick (17B active / 400B total, 128 experts) and LLaMA-4-Scout (17B active / 109B total, 16 experts) \citep{meta2025llama4}, enabling within-family comparison across model scales under the same MoE architecture.
    \item \textbf{Open-weights (Alibaba):} Qwen3-235B-A22B-Instruct (22B active / 235B total) and Qwen3-30B-A3B-Instruct (3B active / 30B total) \citep{qwen2025qwen3}, broadening coverage to a non-Meta open-weights ecosystem and providing a second within-family scale.
\end{itemize}
This selection enables three axes of comparison: closed vs.\ open-source, large vs.\ small scale, and across model families. We note that all four open-weight models adopt Mixture-of-Experts architectures, and discuss the generalizability implications of this selection in Appendix~\ref{app:discussion}.
For the scale-granularity analysis, we additionally evaluate Llama-3-8B-Instruct as a dense-architecture robustness baseline. We therefore distinguish the six primary models used throughout the full experimental design from this seventh model, which appears only in the granularity comparison.

\paragraph{Datasets}
We select three datasets that span distinct cognitive demands:
\begin{itemize}
    \item \textbf{MMLU} \citep{hendrycks2020measuring}: 4-choice knowledge-intensive QA covering 57 subjects. We sample a balanced subset of 1{,}000 instances across difficulty levels.
    \item \textbf{GSM8K} \citep{cobbe2021training}: Multi-step mathematical reasoning. We use the full test set ($N{=}1{,}319$).
    \item \textbf{TruthfulQA} \citep{lin2022truthfulqa}: Probes the model's tendency to produce common misconceptions. We use the multiple-choice variant ($N{=}817$).
\end{itemize}
This selection is motivated by the need to test whether scale effects are stable across tasks that vary in difficulty, answer format, and the degree to which overconfidence is a known failure mode.

\paragraph{Prompting}
We use a standardized zero-shot prompt template: \textit{``Answer the following question. Then, provide your confidence as an integer between [l] and [u], where [l] means no confidence and [u] means absolute certainty.''} The explicit anchoring of the endpoints with semantic labels (``no confidence'' / ``absolute certainty'') is held constant across all scale conditions to isolate the effect of numerical range from verbal framing. Full prompt templates are provided in Appendix~\ref{app:prompts}.

\paragraph{Inference \& Statistical Analysis}
All inferences are performed with temperature $T{=}0$ for deterministic outputs. For each combination of model $\times$ dataset $\times$ scale condition, we collect the full set of available instances ($N \geq 800$), providing sufficient statistical power. To assess the statistical reliability of differences in $meta\text{-}d'$ across scale conditions, we employ a nonparametric bootstrap procedure: we resample the response data 10{,}000 times with replacement and compute 95\% confidence intervals for all metrics. Pairwise comparisons between scale conditions are evaluated via permutation tests ($n{=}10{,}000$), with Bonferroni correction.

\begin{figure*}[t!]
\centering
\includegraphics[width=1.05\linewidth]{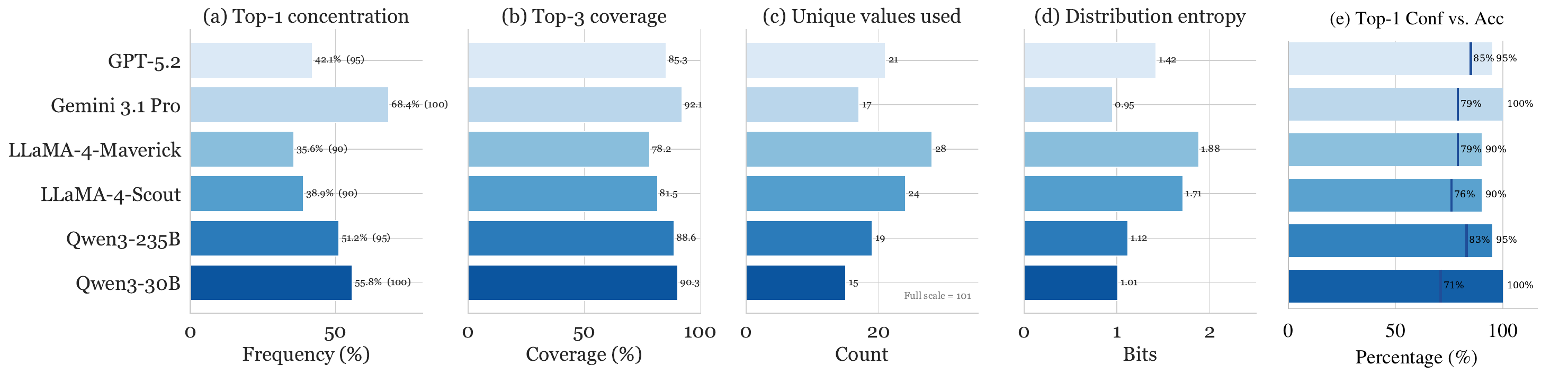}
\caption{Baseline discretization profile under the standard $[0,100]$ confidence scale. Across all six models, confidence reports are highly concentrated on a small set of round-number values. The most frequent value alone accounts for 35.6\%--68.4\% of responses, and the top three values cover 78.2\%--92.1\%. Models use only 15--28 distinct integers out of 101 possible values, revealing severe discretization of the verbalized confidence signal. Conditional accuracy at the Top-1 confidence value, averaged across datasets in (e) show gaps from 10\% to 29\%.}
\label{fig:baseline_hist}
\end{figure*}


\begin{table*}[t]
\centering
\small
\setlength{\tabcolsep}{4pt}
\renewcommand{\arraystretch}{1.05}
\resizebox{\textwidth}{!}{
\begin{tabular}{lccccccccc}
\toprule
& \multicolumn{2}{c}{\textbf{Concentration}} 
& \multicolumn{2}{c}{\textbf{Distribution}} 
& \multicolumn{5}{c}{\textbf{Performance}} \\
\cmidrule(lr){2-3} \cmidrule(lr){4-5} \cmidrule(lr){6-10}
Model & Top-1 & Top-3 (\%) & Unique & $H$ (bits) & Acc & AUROC & ECE $\downarrow$ & $meta\text{-}d'$ & $M_{ratio}$ \\
\midrule
\rowcolor{rowgray}
GPT-5.2          
& 95\;\;(42.1\%)\;\;\ibar{42} 
& 85.3 & 21 & 1.42 & .81 & .88 & .08 & 1.84 & \textbf{0.92} \\
LLaMA-4-Maverick 
& 90\;\;(35.6\%)\;\;\ibar{36} 
& 78.2 & 28 & 1.88 & .76 & .84 & .11 & 1.65 & 0.82 \\
\rowcolor{rowgray}
Qwen3-235B       
& 95\;\;(51.2\%)\;\;\ibar{51} 
& 88.6 & 19 & 1.12 & .79 & .82 & .13 & 1.51 & 0.78 \\
LLaMA-4-Scout    
& 90\;\;(38.9\%)\;\;\ibar{39} 
& 81.5 & 24 & 1.71 & .72 & .80 & .13 & 1.52 & 0.76 \\
\rowcolor{rowgray}
Gemini 3.1 Pro   
& 100\;\;(68.4\%)\;\;\ibar{68} 
& 92.1 & 17 & 0.95 & .78 & .79 & .15 & 1.42 & 0.74 \\
Qwen3-30B        
& 100\;\;(55.8\%)\;\;\ibar{56} 
& 90.3 & 15 & 1.01 & .68 & .74 & .17 & 1.22 & 0.62 \\
\bottomrule
\end{tabular}
}
\caption{Baseline discretization under the standard $[0,100]$ scale, averaged across MMLU, GSM8K, and TruthfulQA. Models are ordered by $M_{ratio}$ (descending). 
\textit{Top-1}: most frequent confidence value (frequency in parentheses; bar proportional to frequency). 
\textit{Top-3}: cumulative coverage of the three most common values. 
\textit{Unique}: number of distinct confidence integers used out of 101 possible values. 
\textit{H}: Shannon entropy in bits. 
\textit{AUROC} for failure prediction. 
\textit{For reference, a uniform distribution over 101 confidence levels would yield 6.66 bits.}}
\label{tab:baseline}
\end{table*}

\begin{table*}[t]
\centering
\small
\begin{tabular}{lcccccccccc}
\toprule
 & \multicolumn{5}{c}{$meta\text{-}d'$ $\uparrow$} & \multicolumn{5}{c}{$M_{ratio}$ $\uparrow$} \\
\cmidrule(lr){2-6} \cmidrule(lr){7-11}
Model & [0,5] & [0,10] & [0,20] & [0,50] & [0,100] & [0,5] & [0,10] & [0,20] & [0,50] & [0,100] \\
\midrule
\rowcolor{rowgray}
GPT-5.2          & 1.68 & 1.75 & \cellcolor{best}\textbf{1.92}$^*$ & 1.80 & 1.84 & 0.84 & 0.90 & \cellcolor{best}\textbf{0.95}$^*$ & 0.91 & 0.92 \\
Gemini 3.1 Pro   & 1.30 & 1.40 & \cellcolor{best}\textbf{1.52}$^*$ & 1.44 & 1.42 & 0.68 & 0.73 & \cellcolor{best}\textbf{0.79}$^*$ & 0.75 & 0.74 \\
\rowcolor{rowgray}
LLaMA-4-Maverick & 1.41 & 1.55 & \cellcolor{best}\textbf{1.73}$^*$ & 1.60 & 1.65 & 0.78 & 0.85 & \cellcolor{best}\textbf{0.89}$^*$ & 0.81 & 0.82 \\
LLaMA-4-Scout    & 1.28 & 1.40 & \cellcolor{best}\textbf{1.55}$^*$ & 1.42 & 1.52 & 0.71 & 0.78 & \cellcolor{best}\textbf{0.83}$^*$ & 0.77 & 0.76 \\
\rowcolor{rowgray}
Qwen3-235B       & 1.39 & 1.53 & \cellcolor{best}\textbf{1.62}$^*$ & 1.55 & 1.51 & 0.72 & 0.79 & \cellcolor{best}\textbf{0.84}$^*$ & 0.80 & 0.78 \\
Qwen3-30B        & 1.08 & 1.20 & \cellcolor{best}\textbf{1.34}$^*$ & 1.24 & 1.22 & 0.55 & 0.61 & \cellcolor{best}\textbf{0.68}$^*$ & 0.63 & 0.62 \\
\rowcolor{rowgray}
Llama-3-8B-Instruct & 1.22 & 1.35 & \cellcolor{best}\textbf{1.48}$^*$ & 1.38 & 1.40 & 0.63 & 0.70 & \cellcolor{best}\textbf{0.79}$^*$ & 0.72 & 0.71 \\
\bottomrule
\end{tabular}
\caption{Effect of scale granularity ($\mathcal{G}$) on metacognitive sensitivity ($meta\text{-}d'$) and efficiency ($M_{ratio}$), averaged across datasets. Llama-3-8B-Instruct is included as a dense architecture baseline.
\colorbox{best}{\strut Shaded cells} indicate the best value per model; $^*$ denotes significant improvement over the $[0,100]$ baseline ($p < 0.05$, permutation test with Bonferroni correction). Confidence values are normalized to $[0,1]$ prior to metric computation.}
\label{tab:granularity}
\end{table*}

\section{Results}
\label{sec:results}

\subsection{Confidence Discretization Under Standard Scales}

We first characterize the distribution of verbalized confidence under the standard $[0, 100]$ scale (Table~\ref{tab:baseline}; distributions visualized in Figure~\ref{fig:baseline_hist}). All six models exhibit severe discretization: models concentrate reports on a sparse set of round-number anchors.

Across models, a single confidence value dominates the distribution. Gemini~3.1~Pro reports exactly 100 on 68.4\% of instances, GPT-5.2 and Qwen3-235B most frequently output 95, and the LLaMA-4 family concentrates on 90. The three most common values together account for more than 78\% of responses for every model. Notably, this concentration is not simply a byproduct of high task accuracy: conditioned on reporting the Top-1 value, accuracy is substantially lower than the value itself would imply: e.g., Gemini~3.1~Pro is correct on only 79\% of the instances where it reports 100, and GPT-5.2 is correct on 85\% of instances where it reports 95. This gap indicates that the dominant confidence value functions as a default token-level output (Figure~\ref{fig:baseline_hist}(e)).
Despite the nominal 101-state scale, models actually use only 15--28 distinct confidence values, leaving most of the scale unused.

Entropy further quantifies this compression. While a uniform distribution over 101 states would yield 6.66 bits, the observed entropy ranges from 0.95 bits (Gemini~3.1~Pro) to 1.88 bits (LLaMA-4-Maverick), indicating highly concentrated confidence distributions.

Baseline metacognitive efficiency ($M_{ratio}$) varies substantially across models, from 0.62 (Qwen3-30B) to 0.92 (GPT-5.2). A consistent pattern emerges: models with lower entropy (i.e., more concentrated confidence distributions) tend to exhibit lower $M_{ratio}$. We examine the causal direction of this relationship through controlled scale manipulation experiments in the following sections. AUROC values remain relatively high across models, indicating that even discretized confidence signals still preserve coarse ranking information between correct and incorrect predictions.


\subsection{Effect of Scale Granularity}

Table~\ref{tab:granularity} reports the effect of scale granularity ($\mathcal{G}$) on metacognitive sensitivity ($meta\text{-}d'$) and efficiency ($M_{ratio}$). Task accuracy remains stable across granularity conditions, confirming that changes in $meta\text{-}d'$ reflect differences in metacognitive signal quality.

\begin{figure}[t!]
\centering
\includegraphics[width=0.95\linewidth]{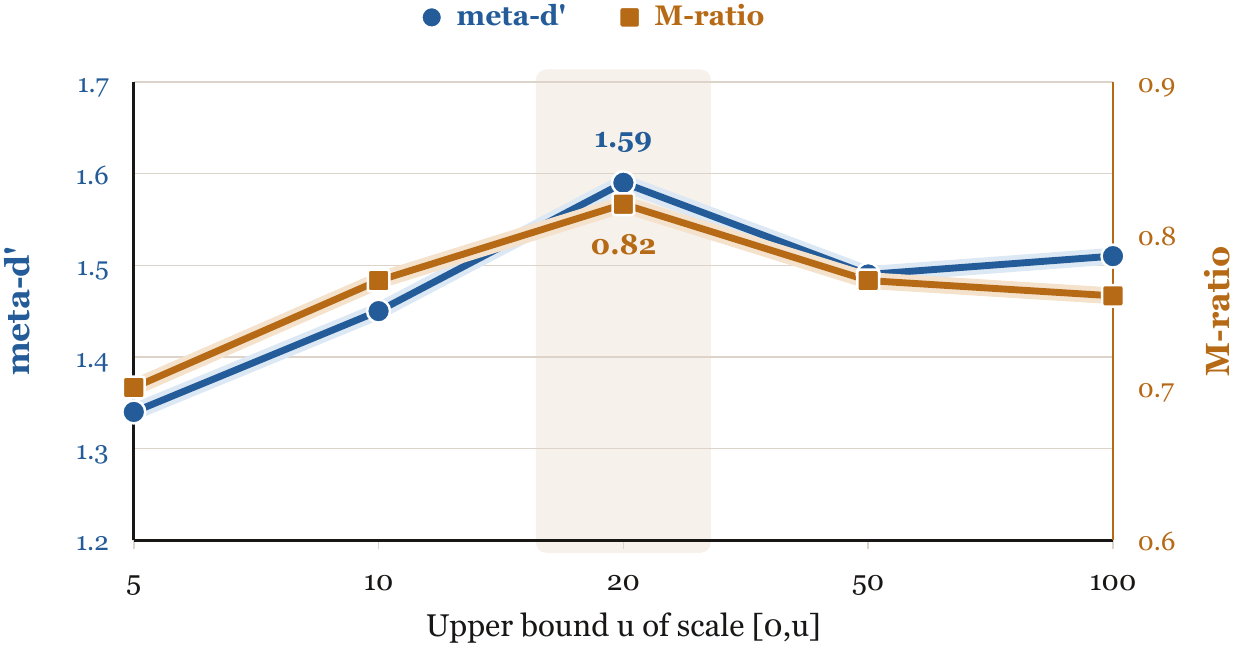}
\caption{Averaged $meta\text{-}d'$ and $M_{ratio}$ across 7 models along scale granularity. Both metrics peak at $[0,20]$ and decline toward both coarser and finer extremes.}
\label{fig:granularity_curve}
\end{figure}

Across the six primary models and the additional dense-architecture baseline, we observe a consistent non-monotonic relationship between scale granularity and metacognitive performance (Figure~\ref{fig:granularity_curve}): averaged across models, both $meta\text{-}d'$ and $M_{ratio}$ peak at $[0,20]$ and decline toward both extremes. At the coarse end, $[0,5]$ yields the lowest $meta\text{-}d'$ for every model, indicating that five categories are insufficient to capture meaningful confidence variation. At the fine end, the standard $[0,100]$ scale does not outperform $[0,20]$ for any model; every model achieves significantly higher $M_{ratio}$ at $[0,20]$ than at $[0,100]$ ($p < 0.05$).

This pattern holds across both closed-source models (e.g.,\ GPT-5.2 $M_{ratio}$ = 0.95 at $[0,20]$ vs.\ 0.92 at $[0,100]$) and open-weight models (e.g.,\ LLaMA-4-Maverick: 0.89 vs.\ 0.82), suggesting that the $[0,20]$ advantage is robust to architectural and training differences. Llama-3-8B-Instruct, a dense architecture model, exhibits the same non-monotonic pattern ($M_{ratio}$ = 0.79 at $[0,20]$ vs.\ 0.71 at $[0,100]$, $p$ < 0.05), confirming that the finding is not specific to Mixture-of-Experts architectures. Per-dataset breakdowns are in Appendix~\ref{app:granularity_full}.

\subsection{Effect of Boundary Shifting}

Table~\ref{tab:boundary} reports the results of the boundary shifting experiment ($\mathcal{B}$), where the upper bound is fixed at 100 while the lower bound is progressively increased. Two patterns deserve attention.

First, moderate boundary shifts have minimal impact. Moving from $[0,100]$ to $[20,100]$ does not produce significant changes in $meta\text{-}d'$ or $M_{ratio}$ for any models, and utilization (\emph{Util.}) increases slightly for several models, suggesting that models can partially adapt to a mildly compressed range. Violation rates under all standard and boundary-shifted conditions are at or near zero (see Table~\ref{tab:violations_full}). 

Second, aggressive boundary shifts lead to consistent degradation. In the $[40,100]$ condition, $M_{ratio}$ decreases for most models, and in the $[60,100]$ condition the drop is universal and substantial. The Utilization metric reveals the mechanism: as the lower bound increases, models fail to redistribute their confidence across the new range and instead cluster more tightly near the upper bound (under the $[60,100]$ condition Qwen3-30B uses only 3.1\% of the available range).

\begin{table}[t]
\centering
\small
\setlength{\tabcolsep}{2pt}
\begin{tabular}{llcccc}
\toprule
 & & \multicolumn{4}{c}{\textbf{Scale Condition}} \\
\cmidrule(lr){3-6}
Model & Metric & [0,100] & [20,100] & [40,100] & [60,100] \\
\midrule
\multirow{3}{*}{GPT-5.2}
& $meta\text{-}d'$ & 1.84 & 1.80 & \cellcolor{degradelight}1.62 & \cellcolor{degradestrong}1.38 \\
& $M_{ratio}$      & 0.92 & 0.90 & \cellcolor{degradelight}0.81$^\dagger$ & \cellcolor{degradestrong}0.69$^\dagger$ \\
& Util.             & 12.4 & \cellcolor{improve}14.1 & 10.8 & \cellcolor{degradestrong}7.2 \\
\midrule
\multirow{3}{*}{\shortstack[l]{Gemini\\3.1 Pro}}
& $meta\text{-}d'$ & 1.42 & 1.45 & 1.28 & \cellcolor{degradestrong}1.12 \\
& $M_{ratio}$      & 0.74 & \cellcolor{improve}0.76 & 0.67 & \cellcolor{degradestrong}0.58$^\dagger$ \\
& Util.             & 6.9  & \cellcolor{improve}8.6 & 6.1 & \cellcolor{degradestrong}4.8 \\
\midrule
\multirow{3}{*}{\shortstack[l]{LLaMA-4\\Maverick}}
& $meta\text{-}d'$ & 1.65 & 1.61 & \cellcolor{degradelight}1.48 & \cellcolor{degradestrong}1.25 \\
& $M_{ratio}$      & 0.82 & 0.80 & \cellcolor{degradelight}0.74$^\dagger$ & \cellcolor{degradestrong}0.63$^\dagger$ \\
& Util.             & 10.5 & \cellcolor{improve}11.8 & 8.9 & \cellcolor{degradestrong}5.6 \\
\midrule
\multirow{3}{*}{\shortstack[l]{LLaMA-4\\Scout}}
& $meta\text{-}d'$ & 1.52 & 1.49 & \cellcolor{degradelight}1.35 & \cellcolor{degradestrong}1.18 \\
& $M_{ratio}$      & 0.76 & 0.74 & \cellcolor{degradelight}0.68$^\dagger$ & \cellcolor{degradestrong}0.59$^\dagger$ \\
& Util.             & 9.2  & \cellcolor{improve}10.1 & 7.5 & \cellcolor{degradestrong}4.9 \\
\midrule
\multirow{3}{*}{\shortstack[l]{Qwen3\\235B}}
& $meta\text{-}d'$ & 1.51 & 1.48 & \cellcolor{degradelight}1.32 & \cellcolor{degradestrong}1.05 \\
& $M_{ratio}$      & 0.78 & 0.76 & \cellcolor{degradelight}0.68$^\dagger$ & \cellcolor{degradestrong}0.55$^\dagger$ \\
& Util.             & 9.1  & 7.4 & 6.2 & \cellcolor{degradestrong}3.9 \\
\midrule
\multirow{3}{*}{\shortstack[l]{Qwen3\\30B}}
& $meta\text{-}d'$ & 1.22 & 1.18 & \cellcolor{degradelight}1.02 & \cellcolor{degradestrong}0.85 \\
& $M_{ratio}$      & 0.62 & 0.60 & \cellcolor{degradelight}0.52$^\dagger$ & \cellcolor{degradestrong}0.43$^\dagger$ \\
& Util.             & 7.8  & 6.5 & 4.8 & \cellcolor{degradestrong}3.1 \\
\bottomrule
\end{tabular}
\caption{
Effect of boundary shifting ($\mathcal{B}$) on metacognitive performance, averaged across datasets.
\textit{Util.}: percentage of the available range covered by the central 90\% of reports (5th--95th percentile), reducing sensitivity to isolated outliers.
Cell shading indicates direction of change relative to $[0,100]$:
\colorbox{improve}{\strut\small improve},
\colorbox{degradelight}{\strut\small moderate$^\dagger$},
\colorbox{degradestrong}{\strut\small severe$^\dagger$}
($^\dagger$: $p < 0.05$, permutation test).
}
\label{tab:boundary}
\vspace{-10pt}
\end{table}

\subsection{Semantic Robustness of Non-standard} 

Tables~\ref{tab:nonstandard_g1} and \ref{tab:nonstandard_g2} present results for the non-standard scale experiment ($\mathcal{N}$), designed to disentangle genuine scale comprehension from round-number.

\paragraph{Group 1: Fixed width, varying bounds}
Under the three fixed-width conditions ($[0,73]$, $[14,86]$, $[7,79]$; Table~\ref{tab:nonstandard_g1}), Round (\textit{round-number preference}) decreases relative to the baseline $[0,100]$ scale but remains high. This indicates that irregular boundaries weaken but do not eliminate round-number clustering. The three Group-1 conditions produce nearly identical results ($M_{ratio}$ within $\pm 0.01$ across models), suggesting that the observed effects are driven by scale irregularity.

\paragraph{Group 2: Diagnostic contrasts}
Table~\ref{tab:nonstandard_g2} reports results for two contrastive conditions. The narrow $[3,38]$ scale produces the strongest disruption: violation rates increase substantially (e.g.,\ up to 21.2\% for Qwen3-30B), indicating that many responses fall outside the instructed range. Metacognitive efficiency also declines ($M_{ratio}$ 0.40--0.72), while Round-Pref remains above 58\% for all models, suggesting persistent preference for multiples of five even when few salient anchors exist.

The minimal pair $[0,97]$ vs.\ $[0,100]$ provides a targeted diagnostic. Despite a difference of only three units in the upper bound, $M_{ratio}$ decreases for every model and Round drops slightly. This consistent shift suggests that the value ``100'' functions as a particularly strong confidence anchor.

These findings suggest that the \textit{discretization patterns observed under standard scales are not merely artifacts of specific numeric boundaries, but reflect persistent anchor-seeking behavior in verbalized confidence generation.}

\begin{table}[t]
\centering
\small
\setlength{\tabcolsep}{2pt}
\begin{tabular}{@{}l l cccc@{}}
\toprule
 & & \multicolumn{4}{c}{\textbf{Scale Condition}} \\
\cmidrule(lr){3-6}
Model & Metric & [0,100] & [0,73] & [14,86] & [7,79] \\
\midrule
\multirow{4}{*}{GPT-5.2}
& Round\,(\%)       & 94.2 & \cellcolor{improve}76.8 & \cellcolor{improve}72.1 & \cellcolor{improve}74.5 \\
& Viol.\,(\%)       & 0.0  & \cellcolor{warnlight}1.8  & \cellcolor{warnlight}2.4  & \cellcolor{warnlight}2.1 \\
& $meta\text{-}d'$  & 1.84 & \cellcolor{degradelight}1.71 & \cellcolor{degradelight}1.68 & \cellcolor{degradelight}1.70 \\
& $M_{ratio}$       & 0.92 & \cellcolor{degradelight}0.85 & \cellcolor{degradelight}0.84 & \cellcolor{degradelight}0.85 \\
\midrule
\multirow{4}{*}{Gemini 3.1}
& Round\,(\%)       & 96.5 & \cellcolor{improve}82.4 & \cellcolor{improve}79.6 & \cellcolor{improve}80.8 \\
& Viol.\,(\%)       & 0.0  & \cellcolor{warnlight}4.1  & \cellcolor{warnlight}5.2  & \cellcolor{warnlight}4.5 \\
& $meta\text{-}d'$  & 1.42 & \cellcolor{degradelight}1.28 & \cellcolor{degradelight}1.24 & \cellcolor{degradelight}1.26 \\
& $M_{ratio}$       & 0.74 & \cellcolor{degradelight}0.67 & \cellcolor{degradelight}0.65 & \cellcolor{degradelight}0.66 \\
\midrule
\multirow{4}{*}{LLaMA-4-M}
& Round\,(\%)       & 88.7 & \cellcolor{improve}71.2 & \cellcolor{improve}68.5 & \cellcolor{improve}69.8 \\
& Viol.\,(\%)       & 0.0  & \cellcolor{warnlight}2.6  & \cellcolor{warnlight}3.1  & \cellcolor{warnlight}2.8 \\
& $meta\text{-}d'$  & 1.65 & \cellcolor{degradelight}1.52 & \cellcolor{degradelight}1.49 & \cellcolor{degradelight}1.50 \\
& $M_{ratio}$       & 0.82 & \cellcolor{degradelight}0.76 & \cellcolor{degradelight}0.74 & \cellcolor{degradelight}0.75 \\
\midrule
\multirow{4}{*}{LLaMA-4-S}
& Round\,(\%)       & 90.1 & \cellcolor{improve}74.8 & \cellcolor{improve}71.9 & \cellcolor{improve}73.2 \\
& Viol.\,(\%)       & 0.0  & \cellcolor{warnlight}3.5  & \cellcolor{warnlight}4.2  & \cellcolor{warnlight}3.8 \\
& $meta\text{-}d'$  & 1.52 & \cellcolor{degradelight}1.38 & \cellcolor{degradelight}1.35 & \cellcolor{degradelight}1.36 \\
& $M_{ratio}$       & 0.76 & \cellcolor{degradelight}0.69 & \cellcolor{degradelight}0.68 & \cellcolor{degradelight}0.68 \\
\midrule
\multirow{4}{*}{Qwen3-235B}
& Round\,(\%)       & 92.1 & \cellcolor{improve}78.4 & \cellcolor{improve}75.2 & \cellcolor{improve}76.8 \\
& Viol.\,(\%)       & 0.0  & \cellcolor{warnlight}3.2  & \cellcolor{warnlight}4.0  & \cellcolor{warnlight}3.5 \\
& $meta\text{-}d'$  & 1.51 & \cellcolor{degradelight}1.38 & \cellcolor{degradelight}1.35 & \cellcolor{degradelight}1.36 \\
& $M_{ratio}$       & 0.78 & \cellcolor{degradelight}0.71 & \cellcolor{degradelight}0.70 & \cellcolor{degradelight}0.70 \\
\midrule
\multirow{4}{*}{Qwen3-30B}
& Round\,(\%)       & 95.3 & \cellcolor{improve}82.6 & \cellcolor{improve}80.1 & \cellcolor{improve}81.2 \\
& Viol.\,(\%)       & 0.0  & \cellcolor{warnlight}5.8  & \cellcolor{warnlight}6.5  & \cellcolor{warnlight}6.1 \\
& $meta\text{-}d'$  & 1.22 & \cellcolor{degradelight}1.05 & \cellcolor{degradelight}1.02 & \cellcolor{degradelight}1.03 \\
& $M_{ratio}$       & 0.62 & \cellcolor{degradelight}0.53 & \cellcolor{degradelight}0.52 & \cellcolor{degradelight}0.52 \\
\bottomrule
\end{tabular}
\caption{
Diagnostics under fixed-width non-standard scales (Group~1, width $\approx$72--73), averaged across datasets.
\textit{Round}: reports on multiples of 5 within $[l,u]$.
\textit{Viol.}: responses outside valid range by $>$5\% of width.
Cell shading:
\colorbox{improve}{\strut\small reduced Round},
\colorbox{degradelight}{\strut\small meta.\ decline},
\colorbox{warnlight}{\strut\small elevated Viol.}
, all relative to $[0,100]$.
}
\label{tab:nonstandard_g1}
\end{table}

\begin{table}[t]
\centering
\small
\setlength{\tabcolsep}{4.5pt}
\begin{tabular}{@{}l l ccc@{}}
\toprule
 & & \multicolumn{3}{c}{\textbf{Scale Condition}} \\
\cmidrule(lr){3-5}
Model & Metric & [0,100] & [3,38] & [0,97] \\
\midrule
\multirow{4}{*}{GPT-5.2}
& Round\,(\%)       & 94.2 & \cellcolor{improve}58.3  & \cellcolor{improve}89.1 \\
& Viol.\,(\%)       & 0.0  & \cellcolor{warnstrong}8.6 & \cellcolor{warnlight}0.5 \\
& $meta\text{-}d'$  & 1.84 & \cellcolor{degradestrong}1.45 & 1.79 \\
& $M_{ratio}$       & 0.92 & \cellcolor{degradestrong}0.72 & 0.90 \\
\midrule
\multirow{4}{*}{Gemini 3.1}
& Round\,(\%)       & 96.5 & \cellcolor{improve}65.2  & \cellcolor{improve}93.8 \\
& Viol.\,(\%)       & 0.0  & \cellcolor{warnstrong}18.3 & \cellcolor{warnlight}0.8 \\
& $meta\text{-}d'$  & 1.42 & \cellcolor{degradestrong}0.95 & 1.38 \\
& $M_{ratio}$       & 0.74 & \cellcolor{degradestrong}0.50 & 0.72 \\
\midrule
\multirow{4}{*}{LLaMA-4-M}
& Round\,(\%)       & 88.7 & \cellcolor{improve}59.4  & \cellcolor{improve}85.3 \\
& Viol.\,(\%)       & 0.0  & \cellcolor{warnstrong}12.9 & \cellcolor{warnlight}0.4 \\
& $meta\text{-}d'$  & 1.65 & \cellcolor{degradestrong}1.28 & 1.61 \\
& $M_{ratio}$       & 0.82 & \cellcolor{degradestrong}0.64 & 0.80 \\
\midrule
\multirow{4}{*}{LLaMA-4-S}
& Round\,(\%)       & 90.1 & \cellcolor{improve}62.1  & \cellcolor{improve}87.5 \\
& Viol.\,(\%)       & 0.0  & \cellcolor{warnstrong}15.4 & \cellcolor{warnlight}0.6 \\
& $meta\text{-}d'$  & 1.52 & \cellcolor{degradestrong}1.14 & 1.48 \\
& $M_{ratio}$       & 0.76 & \cellcolor{degradestrong}0.57 & 0.74 \\
\midrule
\multirow{4}{*}{Qwen3-235B}
& Round\,(\%)       & 92.1 & \cellcolor{improve}62.5  & \cellcolor{improve}89.5 \\
& Viol.\,(\%)       & 0.0  & \cellcolor{warnstrong}14.8 & \cellcolor{warnlight}0.6 \\
& $meta\text{-}d'$  & 1.51 & \cellcolor{degradestrong}1.12 & 1.47 \\
& $M_{ratio}$       & 0.78 & \cellcolor{degradestrong}0.58 & 0.76 \\
\midrule
\multirow{4}{*}{Qwen3-30B}
& Round\,(\%)       & 95.3 & \cellcolor{improve}68.4  & \cellcolor{improve}92.8 \\
& Viol.\,(\%)       & 0.0  & \cellcolor{warnstrong}21.2 & \cellcolor{warnlight}1.2 \\
& $meta\text{-}d'$  & 1.22 & \cellcolor{degradestrong}0.78 & 1.18 \\
& $M_{ratio}$       & 0.62 & \cellcolor{degradestrong}0.40 & 0.60 \\
\bottomrule
\end{tabular}
\caption{
Diagnostics under contrastive non-standard scales (Group~2), averaged across datasets.
$[3,38]$: narrow range, few internal round anchors (stress test).
$[0,97]$: minimal pair with $[0,100]$ (3-unit shift).
Cell shading:
\colorbox{improve}{\strut\small reduced Round},
\colorbox{degradelight}{\strut\small mod.\ decline},
\colorbox{degradestrong}{\strut\small severe decline},
\colorbox{warnstrong}{\strut\small high Viol.}
}
\label{tab:nonstandard_g2}
\end{table}

\subsection{Temperature Robustness}
\label{sec:temperature}

To assess whether the $[0,20]$ advantage is an artifact of greedy decoding, we evaluate GPT-5.2 and LLaMA-4-Maverick on MMLU under stochastic sampling at $T{=}0.3$ and $T{=}1$. Round-number preference decreases with temperature but persists even at $T{=}1$ (GPT-5.2: 88.1\% under $[0,100]$), consistent with anchor-seeking behavior driven by pre-training token frequencies. Critically, $\Delta M_{ratio}$, the improvement of $[0,20]$ over $[0,100]$, remains identical across all three temperature conditions for both models (+0.03 for GPT-5.2, +0.07 for LLaMA-4-Maverick), confirming that the $[0,20]$ advantage is not an artifact of deterministic decoding. Full results are provided in Appendix~\ref{app:temperature}.

\subsection{Prompt Variation}
\label{sec:prompt}

To assess whether the $[0,20]$ advantage depends on our specific prompt wording, we evaluate GPT-5.2 on MMLU under two additional conditions: role prompting (calibrated forecaster persona) and negation framing (reversed endpoint labels). Role prompting produces negligible changes in $M_{ratio}$ relative to baseline, while negation framing introduces a small uniform drop under both scale conditions. In all cases, $\Delta M_{ratio}$ between $[0,20]$ and $[0,100]$ remains +0.03, confirming that the finding is robust to prompt variation. Full results are provided in Appendix~\ref{app:prompt}.

\section{Conclusion}
\label{sec:conclusion}

To our knowledge, we present the first systematic study of confidence scale design as an experimental variable in LLM evaluation. Across six primary models and three datasets, with an additional dense-architecture model used for the granularity robustness analysis, we show that verbalized confidence is severely discretized under the standard $[0,100]$ scale. Within the evaluated settings, a $[0,20]$ scale consistently yields higher metacognitive efficiency, aggressive boundary compression degrades performance, and round-number preference persists even under irregular ranges.
Based on these findings, we recommend three practices: \emph{(i)} use a $[0,20]$ scale as an alternative to the standard $[0,100]$ format; \emph{(ii)} report $meta\text{-}d'$ alongside ECE, since ECE becomes unreliable under highly discretized distributions; and \emph{(iii)} inspect empirical confidence distributions before interpreting calibration metrics. Confidence scale design is often treated as a neutral reporting choice, our results show it is not.

\section*{Project Page and Research Artifacts}
The paper, LaTeX source, figures, and machine-readable citation metadata are publicly available at \url{https://yuyang915.github.io/confidence-scale/}. The corresponding source repository is available at \url{https://github.com/yuyang915/confidence-scale}.

\section*{Limitations and Future Work}
All experiments use greedy decoding ($T{=}0$); temperature robustness is validated in Appendix~\ref{app:temperature}. We use zero-shot prompting without Chain-of-Thought or self-reflection steps; prompt variation results are reported in Appendix~\ref{app:prompt}. Our closed-source models have undisclosed architectures, while open-weight models use Mixture-of-Experts architectures; dense model results are included in Table~\ref{tab:granularity}. We evaluate on three datasets covering knowledge QA, mathematical reasoning, and misconception detection; generalization to open-ended generation tasks is not established. Finally, $meta\text{-}d'$ requires binarizing confidence at a threshold; a continuous Type-2 ROC analysis could provide a more comprehensive characterization and is left to future work.

\section*{Ethics Statement}

This work uses only publicly available benchmark datasets. All model outputs are generated through standard API access or publicly released model weights. We do not anticipate direct negative societal impacts from this research. However, we note that our findings suggest that confidence scores are shaped by scale design rather than solely reflecting genuine model uncertainty. This has implications for deployed systems that present verbalized confidence to end users, since overconfidence driven by scale artifacts could lead to misplaced trust. We encourage practitioners to adopt our recommendations when designing confidence elicitation interfaces.


\bibliographystyle{acl_natbib}
\bibliography{main}

@article{kadavath2022language,
  title={Language models (mostly) know what they know},
  author={Kadavath, Saurav and Conerly, Tom and Askell, Amanda and Henighan, Tom and Drain, Dawn and Perez, Ethan and Schiefer, Nicholas and Hatfield-Dodds, Zac and DasSarma, Nova and Tran-Johnson, Eli and others},
  journal={arXiv preprint arXiv:2207.05221},
  year={2022}
}

@article{lin2022teaching,
  title={Teaching models to express their uncertainty in words},
  author={Lin, Stephanie and Hilton, Jacob and Evans, Owain},
  journal={arXiv preprint arXiv:2205.14334},
  year={2022}
}

@inproceedings{tian2023just,
  title={Just ask for calibration: Strategies for eliciting calibrated confidence scores from language models fine-tuned with human feedback},
  author={Tian, Katherine and Mitchell, Eric and Zhou, Allan and Sharma, Archit and Rafailov, Rafael and Yao, Huaxiu and Finn, Chelsea and Manning, Christopher D},
  booktitle={Proceedings of the 2023 Conference on Empirical Methods in Natural Language Processing},
  pages={5433--5442},
  year={2023}
}

@inproceedings{zhao2021calibrate,
  title={Calibrate before use: Improving few-shot performance of language models},
  author={Zhao, Zihao and Wallace, Eric and Feng, Shi and Klein, Dan and Singh, Sameer},
  booktitle={International conference on machine learning},
  pages={12697--12706},
  year={2021},
  organization={Pmlr}
}

@article{shaki2023cognitive,
  title={Cognitive effects in large language models},
  author={Shaki, Jonathan and Kraus, Sarit and Wooldridge, Michael},
  journal={arXiv preprint arXiv:2308.14337},
  year={2023}
}

@article{pellert2024ai,
  title={Ai psychometrics: Assessing the psychological profiles of large language models through psychometric inventories},
  author={Pellert, Max and Lechner, Clemens M and Wagner, Claudia and Rammstedt, Beatrice and Strohmaier, Markus},
  journal={Perspectives on Psychological Science},
  volume={19},
  number={5},
  pages={808--826},
  year={2024},
  publisher={Sage Publications Sage CA: Los Angeles, CA}
}

@article{lozano2008effect,
  title={Effect of the number of response categories on the reliability and validity of rating scales},
  author={Lozano, Luis M and Garc{\'\i}a-Cueto, Eduardo and Mu{\~n}iz, Jos{\'e}},
  journal={Methodology},
  volume={4},
  number={2},
  pages={73--79},
  year={2008},
  publisher={Hogrefe \& Huber Publishers}
}

@article{tversky1974judgment,
  title={Judgment under Uncertainty: Heuristics and Biases: Biases in judgments reveal some heuristics of thinking under uncertainty.},
  author={Tversky, Amos and Kahneman, Daniel},
  journal={science},
  volume={185},
  number={4157},
  pages={1124--1131},
  year={1974},
  publisher={American association for the advancement of science}
}

@article{harvey1997confidence,
  title={Confidence in judgment},
  author={Harvey, Nigel},
  journal={Trends in cognitive sciences},
  volume={1},
  number={2},
  pages={78--82},
  year={1997},
  publisher={Elsevier}
}

@article{preston2000optimal,
  title={Optimal number of response categories in rating scales: reliability, validity, discriminating power, and respondent preferences},
  author={Preston, Carolyn C and Colman, Andrew M},
  journal={Acta psychologica},
  volume={104},
  number={1},
  pages={1--15},
  year={2000},
  publisher={Elsevier}
}

@article{maniscalco2012signal,
  title={A signal detection theoretic approach for estimating metacognitive sensitivity from confidence ratings},
  author={Maniscalco, Brian and Lau, Hakwan},
  journal={Consciousness and cognition},
  volume={21},
  number={1},
  pages={422--430},
  year={2012},
  publisher={Elsevier}
}

@article{fleming2014measure,
  title={How to measure metacognition},
  author={Fleming, Stephen M and Lau, Hakwan C},
  journal={Frontiers in human neuroscience},
  volume={8},
  pages={82285},
  year={2014},
  publisher={Frontiers}
}

@article{li2024confidence,
  title={Confidence matters: Revisiting intrinsic self-correction capabilities of large language models},
  author={Li, Loka and Chen, Zhenhao and Chen, Guangyi and Zhang, Yixuan and Su, Yusheng and Xing, Eric and Zhang, Kun},
  journal={arXiv preprint arXiv:2402.12563},
  year={2024}
}

@article{xiong2024can,
  title={Can LLMs express their uncertainty},
  author={Xiong, Miao and Hu, Zhiyuan and Lu, Xinyang and Li, Yifei and Fu, Jie and He, Junxian and Hooi, Bryan},
  journal={An empirical evaluation of confidence elicitation in LLMs. arXiv},
  volume={2306},
  year={2024}
}

@inproceedings{guo2017calibration,
  title={On calibration of modern neural networks},
  author={Guo, Chuan and Pleiss, Geoff and Sun, Yu and Weinberger, Kilian Q},
  booktitle={International conference on machine learning},
  pages={1321--1330},
  year={2017},
  organization={PMLR}
}

@inproceedings{zhou2024relying,
  title={Relying on the unreliable: The impact of language models’ reluctance to express uncertainty},
  author={Zhou, Kaitlyn and Hwang, Jena and Ren, Xiang and Sap, Maarten},
  booktitle={Proceedings of the 62nd Annual Meeting of the Association for Computational Linguistics (Volume 1: Long Papers)},
  pages={3623--3643},
  year={2024}
}

@article{wei2022chain,
  title={Chain-of-thought prompting elicits reasoning in large language models},
  author={Wei, Jason and Wang, Xuezhi and Schuurmans, Dale and Bosma, Maarten and Xia, Fei and Chi, Ed and Le, Quoc V and Zhou, Denny and others},
  journal={Advances in neural information processing systems},
  volume={35},
  pages={24824--24837},
  year={2022}
}

@article{wang2022self,
  title={Self-consistency improves chain of thought reasoning in language models},
  author={Wang, Xuezhi and Wei, Jason and Schuurmans, Dale and Le, Quoc and Chi, Ed and Narang, Sharan and Chowdhery, Aakanksha and Zhou, Denny},
  journal={arXiv preprint arXiv:2203.11171},
  year={2022}
}

@book{hautus2021detection,
  title={Detection theory: A user's guide},
  author={Hautus, Michael J and Macmillan, Neil A and Creelman, C Douglas},
  year={2021},
  publisher={Routledge}
}

@article{hautus1995corrections,
  title={Corrections for extreme proportions and their biasing effects on estimated values of $d'$},
  author={Hautus, Michael J},
  journal={Behavior research methods, instruments, \& computers},
  volume={27},
  number={1},
  pages={46--51},
  year={1995},
  publisher={Springer}
}

@article{rausch2023measures,
  title={Measures of metacognitive efficiency across cognitive models of decision confidence.},
  author={Rausch, Manuel and Hellmann, Sebastian and Zehetleitner, Michael},
  journal={Psychological Methods},
  year={2023},
  publisher={American Psychological Association}
}

@inproceedings{wang2024metacognitive,
  title={Metacognitive prompting improves understanding in large language models},
  author={Wang, Yuqing and Zhao, Yun},
  booktitle={Proceedings of the 2024 Conference of the North American Chapter of the Association for Computational Linguistics: Human Language Technologies (Volume 1: Long Papers)},
  pages={1914--1926},
  year={2024}
}

@inproceedings{wang2025decoupling,
  title={Decoupling metacognition from cognition: a framework for quantifying metacognitive ability in LLMs},
  author={Wang, Guoqing and Wu, Wen and Ye, Guangze and Cheng, Zhenxiao and Chen, Xi and Zheng, Hong},
  booktitle={Proceedings of the AAAI Conference on Artificial Intelligence},
  volume={39},
  number={24},
  pages={25353--25361},
  year={2025}
}

@article{galvin2003type,
  title={Type 2 tasks in the theory of signal detectability: Discrimination between correct and incorrect decisions},
  author={Galvin, Susan J and Podd, John V and Drga, Vit and Whitmore, John},
  journal={Psychonomic bulletin \& review},
  volume={10},
  number={4},
  pages={843--876},
  year={2003},
  publisher={Springer}
}

@misc{openai2025gpt5,
  title={Introducing GPT-5},
  author={{OpenAI}},
  year={2025},
  url={https://openai.com/index/introducing-gpt-5/}
}

@misc{google2026gemini3,
  title={Gemini 3: Introducing the latest Gemini AI model from Google},
  author={{Google DeepMind}},
  year={2025},
  url={https://blog.google/products-and-platforms/products/gemini/gemini-3/}
}

@misc{meta2025llama4,
  title={The Llama 4 herd: The beginning of a new era of natively multimodal AI innovation},
  author={{Meta AI}},
  year={2025},
  url={https://ai.meta.com/blog/llama-4-multimodal-intelligence/}
}

@article{qwen2025qwen3,
  title={Qwen3 technical report},
  author={Yang, An and Li, Anfeng and Yang, Baosong and Zhang, Beichen and Hui, Binyuan and Zheng, Bo and Yu, Bowen and Gao, Chang and Huang, Chengen and Lv, Chenxu and others},
  journal={arXiv preprint arXiv:2505.09388},
  year={2025}
}

@inproceedings{nixon2019measuring,
  title={Measuring calibration in deep learning.},
  author={Nixon, Jeremy and Dusenberry, Michael W and Zhang, Linchuan and Jerfel, Ghassen and Tran, Dustin},
  booktitle={CVPR workshops},
  volume={2},
  number={7},
  year={2019}
}

@article{kumar2019verified,
  title={Verified uncertainty calibration},
  author={Kumar, Ananya and Liang, Percy S and Ma, Tengyu},
  journal={Advances in neural information processing systems},
  volume={32},
  year={2019}
}

@article{hendrycks2020measuring,
  title={Measuring massive multitask language understanding},
  author={Hendrycks, Dan and Burns, Collin and Basart, Steven and Zou, Andy and Mazeika, Mantas and Song, Dawn and Steinhardt, Jacob},
  journal={arXiv preprint arXiv:2009.03300},
  year={2020}
}

@article{cobbe2021training,
  title={Training verifiers to solve math word problems},
  author={Cobbe, Karl and Kosaraju, Vineet and Bavarian, Mohammad and Chen, Mark and Jun, Heewoo and Kaiser, Lukasz and Plappert, Matthias and Tworek, Jerry and Hilton, Jacob and Nakano, Reiichiro and others},
  journal={arXiv preprint arXiv:2110.14168},
  year={2021}
}

@inproceedings{lin2022truthfulqa,
  title={Truthfulqa: Measuring how models mimic human falsehoods},
  author={Lin, Stephanie and Hilton, Jacob and Evans, Owain},
  booktitle={Proceedings of the 60th annual meeting of the association for computational linguistics (volume 1: long papers)},
  pages={3214--3252},
  year={2022}
}

\clearpage
\appendix
\appendix

\section*{Appendix Overview}
\label{app:overview}
\begin{itemize}[nosep]
    \item[\ref{app:related_work}] Related Work
    \item[\ref{app:ece_bins}] Sensitivity of ECE to Bin Count and Binning Strategy
    \item[\ref{app:prompts}] Prompt Templates
    \item[\ref{app:full_results}] Full Experimental Results
    \begin{itemize}[nosep]
        \item[\ref{app:baseline_per_dataset}] Per-Dataset Baseline Statistics
        \item[\ref{app:gap_per_dataset}] Per-Dataset Conditional Accuracy Gap
        \item[\ref{app:granularity_full}] Granularity: ECE and Per-Dataset Breakdown
        \item[\ref{app:granularity_avg}] Averaged Granularity Curve: Underlying Data
        \item[\ref{app:accuracy_stability}] Task Accuracy Across Granularity Conditions
        \item[\ref{app:boundary_full}] Boundary Shifting: Complete Results
        \item[\ref{app:nonstandard_full}] Non-standard Scales: Complete Results
        \item[\ref{app:confidence_intervals}] Bootstrap Confidence Intervals for Key Metrics
        \item[\ref{app:violations}] Out-of-Range Violation Analysis
    \end{itemize}
    \item[\ref{app:temperature}] Temperature Robustness
    \item[\ref{app:prompt}] Prompt Variation
    \item[\ref{app:logit}] Logit-Level Analysis
    \item[\ref{app:discussion}] Discussion
\end{itemize}

\newpage

\section{Related Work}
\label{app:related_work}

\subsection{Verbalized Confidence and Calibration}

As LLMs increasingly became available through black-box commercial APIs, research on uncertainty estimation shifted from logit-based methods \citep{kadavath2022language} toward verbalized confidence, where models are prompted to express certainty as a numerical score \citep{lin2022teaching, xiong2024can}. Substantial effort has focused on improving confidence elicitation through prompting and inference strategies, including Chain-of-Thought prompting \citep{wei2022chain}, metacognitive prompting \citep{wang2024metacognitive}, self-consistency aggregation \citep{wang2022self}, and techniques tailored to RLHF-aligned models \citep{tian2023just}. Evaluation has likewise centered on calibration metrics, predominantly Expected Calibration Error \citep[ECE:][]{guo2017calibration}. Yet this literature largely treats the confidence scale as a fixed backdrop, assuming that a 0--100 scale faithfully conveys whatever metacognitive signal the model possesses.

However, we find that LLMs do not use the 0--100 scale as a continuous spectrum but cluster outputs around a small set of round-number anchors. Then, when models concentrate the vast majority of their reports on two or three values, ECE bins become sparse or empty across most of the range, making calibration estimates unreliable. This \textit{discretization bias} and its dependence on scale design have not been systematically studied.

\subsection{Scale Design in Psychometrics}
In human subject research, scale design is known to be non-neutral: the number of response categories modulates both reliability and validity \citep{lozano2008effect, preston2000optimal}, and scale boundaries bias judgments toward specific anchors \citep{tversky1974judgment, harvey1997confidence}. While such sensitivities are well-documented for human respondents, it remains unclear whether they apply to LLMs, whose outputs are driven by token-level distributions rather than human cognition. LLMs may exhibit analogous effects, or their biases may reflect an additional mechanism not present in human respondents: the statistical prevalence of certain numerical tokens in pre-training corpora \citep{zhao2021calibrate, shaki2023cognitive}.

\subsection{Metacognitive Sensitivity and Signal Detection Theory}
Evaluating metacognition requires distinguishing Type-1 performance (task accuracy) from Type-2 sensitivity (the ability to monitor that accuracy) \citep{galvin2003type}. Signal Detection Theory (SDT) provides a principled framework for this decomposition \citep{hautus2021detection}. The $meta\text{-}d'$ metric \citep{maniscalco2012signal} quantifies metacognitive sensitivity by measuring how well confidence ratings distinguish correct from incorrect answers, independently of response bias \citep{fleming2014measure, rausch2023measures}. Standard SDT corrections for extreme proportions ensure robust estimation under skewed response distributions~\citep{hautus1995corrections}. Though recent NLP work has begun disentangling metacognitive ability from task performance \citep{wang2025decoupling}, no study has used SDT-based metrics to examine how confidence scale design influences LLM metacognition. By applying $meta\text{-}d'$ across controlled confidence scales, we address this gap.

\section{Sensitivity of ECE to Bin Count and Binning Strategy}
\label{app:ece_bins}

A known limitation of Expected Calibration Error is its dependence on the number of bins $B$ and the binning strategy \citep{nixon2019measuring, kumar2019verified}. This sensitivity is particularly acute in our setting, where confidence distributions are heavily discretized---the majority of probability mass is concentrated on two or three values (e.g., 95 and 100 under the $[0,100]$ scale). Under such distributions, most equal-width bins are empty or near-empty, and small changes in $B$ can shift samples across bin boundaries, producing unstable ECE estimates.

\subsection{Effect of $B$ on ECE Estimates}
\label{app:ece_b_effect}

\begin{figure*}[t!]
    \centering
    \includegraphics[width=0.98\linewidth]{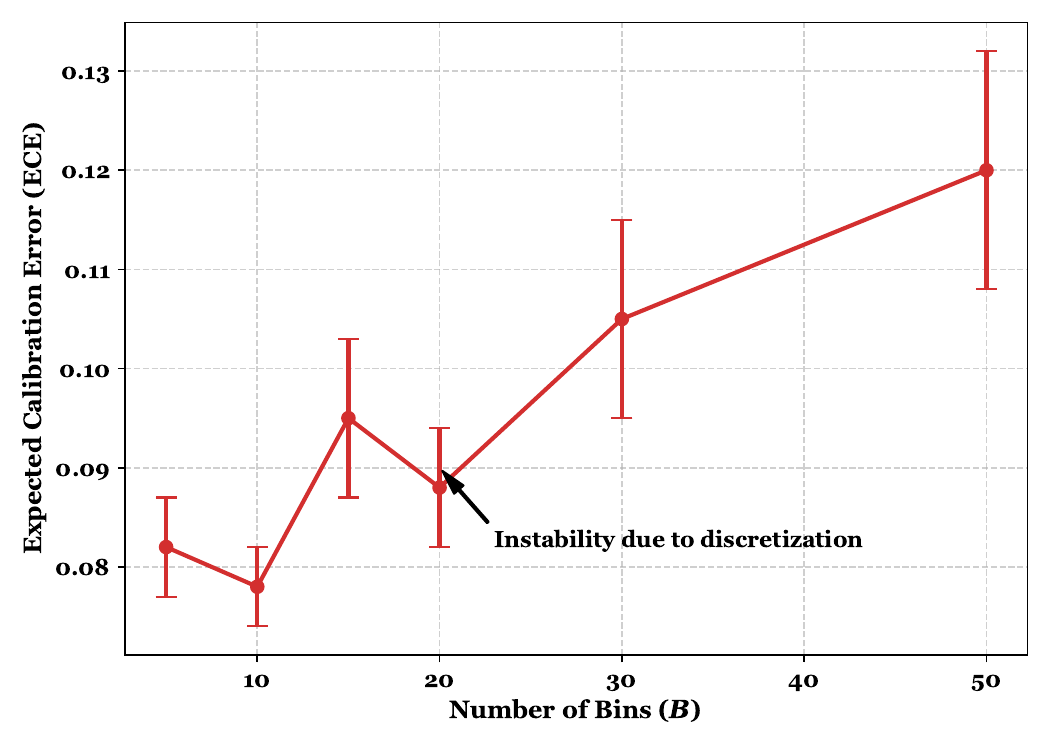}
    \caption{Sensitivity of Expected Calibration Error (ECE) to the number of bins $B$ under discretized confidence distributions. When confidence values concentrate on a small set of anchors, small changes in $B$ can shift observations across bin boundaries, producing unstable ECE estimates.}
    \label{fig:ece_sensitivity}
\end{figure*}

Figure~\ref{fig:ece_sensitivity} shows how ECE varies as a function of $B \in \{5, 10, 15, 20, 30, 50\}$ for a representative model--dataset pair (GPT-5.2 on MMLU) under the standard $[0,100]$ scale. ECE fluctuates noticeably across bin counts, with relative differences of up to approximately 13\% between $B{=}10$ and $B{=}20$. This instability arises because the model's confidence reports are highly discretized and concentrated on a small set of values (e.g., 90, 95, and 100). After normalization to the $[0,1]$ interval, most probability mass falls within a narrow region of the scale, causing bin membership to be highly sensitive to boundary placement. As a result, small changes in $B$ can reassign a substantial number of observations across bins, producing noticeable variation in the estimated calibration error.

\subsection{Equal-Width vs.\ Equal-Mass Binning}
\label{app:binning_strategy}

The standard ECE formulation uses \textit{equal-width} bins, which partition $[0,1]$ into $B$ intervals of size $1/B$. Under discretized confidence, this produces a highly uneven distribution of samples across bins. An alternative is \textit{equal-mass} (or \textit{quantile}) binning, which assigns approximately $N/B$ samples to each bin, ensuring that no bin is empty \citep{nixon2019measuring}.

Table~\ref{tab:ece_binning} reports ECE under both strategies with $B{=}10$ for representative model--scale combinations on MMLU. We highlight two patterns. First, under the standard $[0,100]$ scale, the discrepancy between equal-width and equal-mass ECE is substantial, reaching up to $0.037$ absolute points (e.g., for LLaMA-4-Maverick). This confirms that binning artifacts can dominate the calibration estimate when confidence values are concentrated on a small set of anchors. Second, under the coarser $[0,20]$ scale, the discrepancy shrinks considerably (typically below $0.01$), indicating that the distribution of confidence values becomes more evenly spread across bins. Importantly, these differences arise despite identical model predictions and confidence reports, indicating that the observed variation is purely an artifact of the binning procedure. This result supports the interpretation that scale design directly shapes the empirical confidence distribution and, consequently, the stability of calibration metrics.

\begin{table}[h]
\centering
\small
\caption{Comparison of ECE under \textbf{Equal-Width} vs. \textbf{Equal-Mass} binning ($B=10$) on MMLU. Large $|\Delta|$ indicates metrics driven by discretization artifacts.}
\label{tab:ece_binning}
\begin{tabular}{llccc}
\toprule
Model & Scale & Equal-Width & Equal-Mass & $|\Delta|$ \\
\midrule
GPT-5.2          & [0, 100] & 0.082 & 0.051 & 0.031 \\
                 & [0, 20]  & 0.054 & 0.048 & 0.006 \\
\midrule
LLaMA-4-Mav.     & [0, 100] & 0.115 & 0.078 & 0.037 \\
                 & [0, 20]  & 0.082 & 0.075 & 0.007 \\
\bottomrule
\end{tabular}
\end{table}

\subsection{Complete ECE Results Across Bin Counts}
\label{app:ece_full}

For completeness, Table~\ref{tab:ece_full} reports ECE with $B \in \{10, 15, 20\}$ for all experimental conditions (model $\times$ dataset $\times$ scale). Our main findings are robust to the choice of $B$: although absolute ECE values vary as the number of bins changes, the relative ordering of scale conditions remains stable across all tested bin counts. In particular, the $[0,20]$ scale consistently yields the lowest ECE values, followed by $[0,50]$, while the standard $[0,100]$ scale produces the highest ECE across all settings.

This stability in ranking indicates that our conclusions about scale design are not driven by a specific choice of bin count. Instead, the differences across scales reflect systematic changes in the empirical confidence distribution. At the same time, the noticeable variation in absolute ECE values across $B$ further illustrates the sensitivity of bin-based calibration metrics under discretized confidence distributions, reinforcing the motivation for using $meta\text{-}d'$ as a complementary, distribution-agnostic measure of metacognitive sensitivity.

\begin{table*}[h]
\centering
\small
\caption{ECE across experimental conditions with $B \in \{10, 15, 20\}$. Ranking of scales is stable ($[0,20] < [0,50] < [0,100]$) across different $B$ values.}
\label{tab:ece_full}
\begin{tabular}{lccc|ccc}
\toprule
 & \multicolumn{3}{c}{GPT-5.2 (MMLU)} & \multicolumn{3}{c}{LLaMA-4-Mav (MMLU)} \\
Scale & $B=10$ & $B=15$ & $B=20$ & $B=10$ & $B=15$ & $B=20$ \\
\midrule
$[0,20]$  & \textbf{0.051} & \textbf{0.054} & \textbf{0.058} & \textbf{0.082} & \textbf{0.085} & \textbf{0.089} \\
$[0,50]$  & 0.078 & 0.082 & 0.085 & 0.105 & 0.112 & 0.118 \\
$[0,100]$ & 0.082 & 0.095 & 0.102 & 0.115 & 0.132 & 0.145 \\
\bottomrule
\end{tabular}
\end{table*}

\subsection{Confidence Distribution Histograms}
\label{app:conf_distributions}

\begin{figure*}[h]
    \centering
    \includegraphics[width=1.1\linewidth]{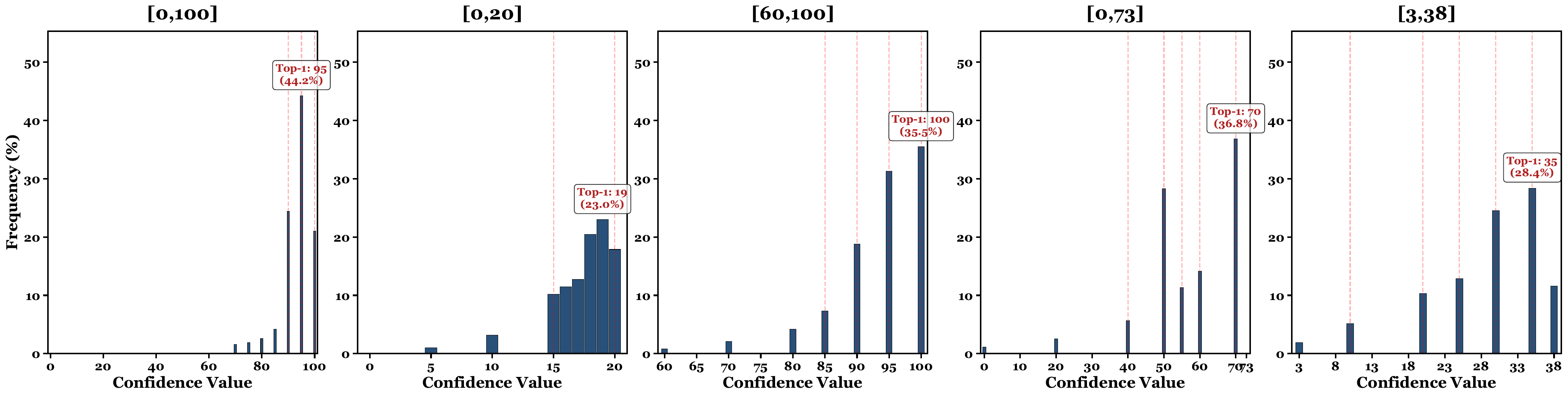}
    \caption{Distribution of raw confidence values across representative scale conditions for GPT-5.2, aggregated across datasets. Under the standard $[0,100]$ scale, confidence reports concentrate on a small set of round-number anchors, whereas alternative scale designs produce qualitatively different distributional patterns. Dashed vertical lines indicate the most frequent round-number anchors within each range.}
    \label{fig:conf_histograms}
\end{figure*}

\begin{figure*}[h]
    \centering
    \includegraphics[width=1.1\linewidth]{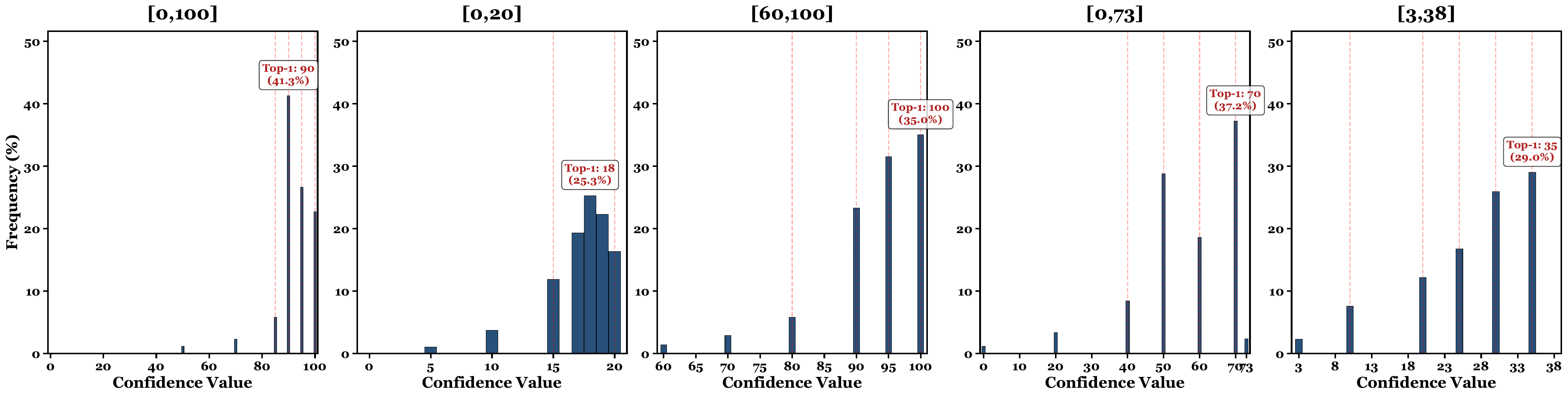}
    \caption{Distribution of raw confidence values across representative scale conditions for LLaMA-4-Maverick, aggregated across datasets. Similar to GPT-5.2, the standard $[0,100]$ scale produces strong clustering on round-number anchors, while coarser, shifted, and non-standard scales alter the distributional pattern. Dashed vertical lines indicate the most frequent round-number anchors within each range.}
    \label{fig:conf1_histograms}
\end{figure*}

Figures~\ref{fig:conf_histograms} and \ref{fig:conf1_histograms} visualize the raw (pre-normalization) confidence distributions for each scale condition, aggregated across datasets. These histograms directly illustrate the discretization phenomenon discussed in Section~\ref{sec:results} and motivate our use of $meta\text{-}d'$ as a distribution-agnostic alternative to ECE.

These visualizations serve three purposes: (1) they provide direct evidence of discretization as a function of scale design; (2) they explain \textit{why} ECE estimates are unstable---when 80\%+ of samples fall in a single bin, the metric is dominated by that bin's accuracy-confidence gap; and (3) they visually preview the key result that coarser and non-standard scales can reshape the confidence distribution in ways that are informative for analysis in the main text.

\section{Prompt Templates}
\label{app:prompts}

All prompts follow a standardized two-part structure: a \textit{task instruction} that presents the question and elicits an answer, followed by a \textit{confidence instruction} that specifies the scale and requests a numerical self-assessment. The task instruction is held constant across all scale conditions; only the confidence instruction varies. This design isolates the effect of scale design from any confounds introduced by prompt wording.

\subsection{General Template}

\begin{quote}
\small
\texttt{[TASK INSTRUCTION]}\\[4pt]
\texttt{Answer the following question.}\\
\texttt{\{question\}}\\
\texttt{Choices: \{choices\}}\\[4pt]
\texttt{[CONFIDENCE INSTRUCTION]}\\
\texttt{After providing your answer, rate your confidence as an integer between \{l\} and \{u\}, where \{l\} means no confidence at all and \{u\} means absolute certainty.}\\[4pt]
\texttt{[OUTPUT FORMAT]}\\
\texttt{Respond in the following format only:}\\
\texttt{Answer: <your answer>}\\
\texttt{Confidence: <integer between \{l\} and \{u\}>}
\end{quote}

\noindent The explicit anchoring of the endpoints with semantic labels (``no confidence at all'' / ``absolute certainty'') is held constant across all conditions. The strict output format minimizes post-processing ambiguity and discourages the model from hedging or elaborating on its confidence.

For GSM8K (free-form numerical answers), the task instruction is adapted as follows:

\begin{quote}
\small
\texttt{Solve the following math problem step by step.}\\
\texttt{\{question\}}\\[4pt]
\texttt{Respond in the following format only:}\\
\texttt{Solution: <your step-by-step solution>}\\
\texttt{Final Answer: <numerical answer>}\\
\texttt{Confidence: <integer between \{l\} and \{u\}>}
\end{quote}

\subsection{Scale-Specific Confidence Instructions}

Below we list the exact confidence instruction for each scale condition. All other parts of the prompt remain identical.

\subsubsection{Granularity Conditions ($\mathcal{G}$)}

\begin{quote}
\small
\textbf{[0, 5]:} \texttt{Rate your confidence as an integer between 0 and 5, where 0 means no confidence at all and 5 means absolute certainty.}\\[6pt]
\textbf{[0, 10]:} \texttt{Rate your confidence as an integer between 0 and 10, where 0 means no confidence at all and 10 means absolute certainty.}\\[6pt]
\textbf{[0, 20]:} \texttt{Rate your confidence as an integer between 0 and 20, where 0 means no confidence at all and 20 means absolute certainty.}\\[6pt]
\textbf{[0, 50]:} \texttt{Rate your confidence as an integer between 0 and 50, where 0 means no confidence at all and 50 means absolute certainty.}\\[6pt]
\textbf{[0, 100]:} \texttt{Rate your confidence as an integer between 0 and 100, where 0 means no confidence at all and 100 means absolute certainty.}
\end{quote}

\subsubsection{Boundary Shifting Conditions ($\mathcal{B}$)}

\begin{quote}
\small
\textbf{[0, 100]:} (Same as granularity baseline above.)\\[6pt]
\textbf{[20, 100]:} \texttt{Rate your confidence as an integer between 20 and 100, where 20 means no confidence at all and 100 means absolute certainty.}\\[6pt]
\textbf{[40, 100]:} \texttt{Rate your confidence as an integer between 40 and 100, where 40 means no confidence at all and 100 means absolute certainty.}\\[6pt]
\textbf{[60, 100]:} \texttt{Rate your confidence as an integer between 60 and 100, where 60 means no confidence at all and 100 means absolute certainty.}
\end{quote}

\subsubsection{Non-standard Range Conditions ($\mathcal{N}$)}

\paragraph{Group 1: Fixed width ($\approx$\,72--73 units), varying bounds.}

\begin{quote}
\small
\textbf{[0, 73]:} \texttt{Rate your confidence as an integer between 0 and 73, where 0 means no confidence at all and 73 means absolute certainty.}\\[6pt]
\textbf{[14, 86]:} \texttt{Rate your confidence as an integer between 14 and 86, where 14 means no confidence at all and 86 means absolute certainty.}\\[6pt]
\textbf{[7, 79]:} \texttt{Rate your confidence as an integer between 7 and 79, where 7 means no confidence at all and 79 means absolute certainty.}
\end{quote}

\paragraph{Group 2: Varying width, diagnostic contrasts.}

\begin{quote}
\small
\textbf{[3, 38]:} \texttt{Rate your confidence as an integer between 3 and 38, where 3 means no confidence at all and 38 means absolute certainty.}\\[6pt]
\textbf{[0, 97]:} \texttt{Rate your confidence as an integer between 0 and 97, where 0 means no confidence at all and 97 means absolute certainty.}
\end{quote}

\subsection{Design Rationale and Controls}

Several design choices warrant explicit justification:

\paragraph{Semantic anchoring.} Every confidence instruction anchors the lower bound to ``no confidence at all'' and the upper bound to ``absolute certainty,'' regardless of the numerical values. This ensures that any observed differences across scale conditions are attributable to the numerical range itself, not to differences in verbal framing. Without this control, a model might interpret ``20'' in a $[20, 100]$ scale as moderate confidence rather than as the minimum, confounding boundary shifting effects with semantic reinterpretation.

\paragraph{Integer constraint.} We explicitly instruct the model to provide an integer. This prevents the model from reporting values like ``95.5'' or ``around 90,'' which would introduce parsing ambiguity. In pilot experiments, omitting this constraint led to approximately 8\% of responses containing non-integer or range-valued confidence reports (e.g., ``85--90'').

\paragraph{Output format enforcement.} The strict \texttt{Answer: / Confidence:} format serves two purposes: (1) it enables reliable automated extraction of confidence values, and (2) it discourages the model from embedding caveats or qualifications that might implicitly modulate the confidence signal (e.g., ``I'm fairly confident, so 85''). Responses that do not conform to the specified format are excluded from analysis. Violation rates for each scale condition are reported in Table~\ref{tab:violations_full}.

\paragraph{No Chain-of-Thought for confidence.} We deliberately do not prompt the model to ``explain your confidence'' or ``think step by step about your certainty.'' While such prompts might improve calibration \citep{xiong2024can, tian2023just}, they would introduce an additional variable that interacts with scale design. Our goal is to measure the raw effect of scale manipulation on the default confidence signal; the interaction between prompting strategies and scale design is left for future work.

\section{Full Experimental Results}
\label{app:full_results}

This appendix provides the complete experimental results underlying the analyses in Section~\ref{sec:results}. Main-text tables report averages across datasets and, for selected results, only representative subsets due to space constraints. Here we provide per-dataset baselines, per-dataset conditional accuracy gaps, per-dataset granularity breakdowns, full-model comparisons, and statistical confidence intervals.

\subsection{Per-Dataset Baseline Statistics}
\label{app:baseline_per_dataset}

Tables~\ref{tab:baseline_mmlu}, \ref{tab:baseline_gsm8k}, and \ref{tab:baseline_truthfulqa} report the baseline discretization statistics from Section~\ref{sec:results} separately for each dataset. This decomposition verifies that the severe confidence discretization observed in the averaged results is consistent across knowledge-intensive QA, mathematical reasoning, and misconception-probing tasks.

\begin{table*}[h]
\centering
\caption{Baseline discretization under $[0,100]$ --- \textbf{MMLU} (knowledge-intensive QA).}
\label{tab:baseline_mmlu}
\small
\begin{tabular}{lcccccccc}
\toprule
Model & Top-1 & Top-3 (\%) & Unique & $H$ (bits) & Acc & ECE $\downarrow$ & $meta\text{-}d'$ & $M_{ratio}$ \\
\midrule
GPT-5.2          & 95  & 83.2 & 23 & 1.48 & 0.85 & 0.07 & 1.95 & 0.94 \\
Gemini 3.1 Pro   & 100 & 91.5 & 18 & 0.98 & 0.82 & 0.14 & 1.48 & 0.75 \\
LLaMA-4-Maverick & 90  & 76.5 & 30 & 1.95 & 0.80 & 0.10 & 1.72 & 0.84 \\
LLaMA-4-Scout    & 90  & 80.1 & 25 & 1.78 & 0.75 & 0.12 & 1.58 & 0.78 \\
Qwen3-235B       & 95  & 87.2 & 20 & 1.15 & 0.83 & 0.12 & 1.58 & 0.79 \\
Qwen3-30B        & 100 & 89.5 & 16 & 1.05 & 0.72 & 0.16 & 1.28 & 0.64 \\
\bottomrule
\end{tabular}
\end{table*}

\begin{table*}[h]
\centering
\caption{Baseline discretization under $[0,100]$ --- \textbf{GSM8K} (mathematical reasoning).}
\label{tab:baseline_gsm8k}
\small
\begin{tabular}{lcccccccc}
\toprule
Model & Top-1 & Top-3 (\%) & Unique & $H$ (bits) & Acc & ECE $\downarrow$ & $meta\text{-}d'$ & $M_{ratio}$ \\
\midrule
GPT-5.2          & 95  & 85.8 & 19 & 1.35 & 0.78 & 0.09 & 1.75 & 0.90 \\
Gemini 3.1 Pro   & 100 & 93.4 & 14 & 0.88 & 0.75 & 0.17 & 1.35 & 0.72 \\
LLaMA-4-Maverick & 90  & 80.2 & 24 & 1.75 & 0.72 & 0.12 & 1.55 & 0.80 \\
LLaMA-4-Scout    & 90  & 83.4 & 22 & 1.62 & 0.68 & 0.14 & 1.45 & 0.74 \\
Qwen3-235B       & 95  & 90.1 & 17 & 1.05 & 0.75 & 0.15 & 1.42 & 0.76 \\
Qwen3-30B        & 100 & 92.5 & 13 & 0.92 & 0.62 & 0.19 & 1.15 & 0.60 \\
\bottomrule
\end{tabular}
\end{table*}

\begin{table*}[h]
\centering
\caption{Baseline discretization under $[0,100]$ --- \textbf{TruthfulQA} (misconception probing).}
\label{tab:baseline_truthfulqa}
\small
\begin{tabular}{lcccccccc}
\toprule
Model & Top-1 & Top-3 (\%) & Unique & $H$ (bits) & Acc & ECE $\downarrow$ & $meta\text{-}d'$ & $M_{ratio}$ \\
\midrule
GPT-5.2          & 95  & 81.5 & 20 & 1.38 & 0.74 & 0.10 & 1.68 & 0.90 \\
Gemini 3.1 Pro   & 100 & 93.2 & 15 & 0.92 & 0.70 & 0.18 & 1.32 & 0.70 \\
LLaMA-4-Maverick & 90  & 78.4 & 26 & 1.82 & 0.71 & 0.13 & 1.58 & 0.81 \\
LLaMA-4-Scout    & 90  & 82.3 & 23 & 1.65 & 0.65 & 0.15 & 1.45 & 0.75 \\
Qwen3-235B       & 95  & 88.9 & 18 & 1.10 & 0.72 & 0.15 & 1.48 & 0.77 \\
Qwen3-30B        & 100 & 91.2 & 14 & 0.98 & 0.58 & 0.21 & 1.10 & 0.58 \\
\bottomrule
\end{tabular}
\end{table*}

\subsection{Per-Dataset Conditional Accuracy Gap}
\label{app:gap_per_dataset}

Tables~\ref{tab:gap_mmlu}, \ref{tab:gap_gsm8k}, and \ref{tab:gap_truthfulqa} report the Top-1 conditional accuracy gap (cf.\ Figure~\ref{fig:baseline_hist}(e) in the main text) separately for each dataset, verifying that the overconfidence pattern at the dominant confidence anchor is not driven by a single task type. Averaging each model's per-dataset gap recovers the values reported in the main text (e.g., GPT-5.2: $(6+10+14)/3 = 10$ pts).

\begin{table}[h]
\centering
\small
\setlength{\tabcolsep}{0.6pt}
\renewcommand{\arraystretch}{1.05}
\caption{Conditional accuracy gap under $[0,100]$ --- \textbf{MMLU}.}
\label{tab:gap_mmlu}
\begin{tabular}{lccccc}
\toprule
Model & Top-1 & \% Rep. & Rep.\ Conf. & Cond.\ Acc. & Gap \\
\midrule
\rowcolor{rowgray}
GPT-5.2          & 95  & 40.8\% & 0.95 & 0.89 & \cellcolor{degradelight}6 pts \\
LLaMA-4-Maverick & 90  & 33.9\% & 0.90 & 0.83 & \cellcolor{degradelight}7 pts \\
\rowcolor{rowgray}
Qwen3-235B       & 95  & 49.5\% & 0.95 & 0.87 & \cellcolor{degradelight}8 pts \\
LLaMA-4-Scout    & 90  & 37.2\% & 0.90 & 0.80 & \cellcolor{degradelight}10 pts \\
\rowcolor{rowgray}
Gemini 3.1 Pro   & 100 & 66.1\% & 1.00 & 0.83 & \cellcolor{degradestrong}17 pts \\
Qwen3-30B        & 100 & 53.4\% & 1.00 & 0.75 & \cellcolor{degradestrong}25 pts \\
\bottomrule
\end{tabular}
\end{table}

\begin{table}[h]
\centering
\small
\setlength{\tabcolsep}{0.6pt}
\renewcommand{\arraystretch}{1.05}
\caption{Conditional accuracy gap under $[0,100]$ --- \textbf{GSM8K}.}
\label{tab:gap_gsm8k}
\begin{tabular}{lccccc}
\toprule
Model & Top-1 & \% Rep. & Rep.\ Conf. & Cond.\ Acc. & Gap \\
\midrule
\rowcolor{rowgray}
GPT-5.2          & 95  & 43.1\% & 0.95 & 0.85 & \cellcolor{degradelight}10 pts \\
LLaMA-4-Maverick & 90  & 36.4\% & 0.90 & 0.79 & \cellcolor{degradelight}11 pts \\
\rowcolor{rowgray}
Qwen3-235B       & 95  & 52.0\% & 0.95 & 0.83 & \cellcolor{degradelight}12 pts \\
LLaMA-4-Scout    & 90  & 39.5\% & 0.90 & 0.76 & \cellcolor{degradelight}14 pts \\
\rowcolor{rowgray}
Gemini 3.1 Pro   & 100 & 69.2\% & 1.00 & 0.79 & \cellcolor{degradestrong}21 pts \\
Qwen3-30B        & 100 & 56.8\% & 1.00 & 0.71 & \cellcolor{degradestrong}29 pts \\
\bottomrule
\end{tabular}
\end{table}

\begin{table}[h]
\centering
\small
\setlength{\tabcolsep}{0.6pt}
\renewcommand{\arraystretch}{1.05}
\caption{Conditional accuracy gap under $[0,100]$ --- \textbf{TruthfulQA}.}
\label{tab:gap_truthfulqa}
\begin{tabular}{lccccc}
\toprule
Model & Top-1 & \% Rep. & Rep.\ Conf. & Cond.\ Acc. & Gap \\
\midrule
\rowcolor{rowgray}
GPT-5.2          & 95  & 42.4\% & 0.95 & 0.81 & \cellcolor{degradelight}14 pts \\
LLaMA-4-Maverick & 90  & 36.5\% & 0.90 & 0.75 & \cellcolor{degradelight}15 pts \\
\rowcolor{rowgray}
Qwen3-235B       & 95  & 52.1\% & 0.95 & 0.79 & \cellcolor{degradestrong}16 pts \\
LLaMA-4-Scout    & 90  & 39.9\% & 0.90 & 0.72 & \cellcolor{degradestrong}18 pts \\
\rowcolor{rowgray}
Gemini 3.1 Pro   & 100 & 69.8\% & 1.00 & 0.75 & \cellcolor{degradestrong}25 pts \\
Qwen3-30B        & 100 & 57.2\% & 1.00 & 0.67 & \cellcolor{degradestrong}33 pts \\
\bottomrule
\end{tabular}
\end{table}

\subsection{Granularity: ECE and Per-Dataset Breakdown}
\label{app:granularity_full}

Table~\ref{tab:granularity_ece} reports ECE across all granularity conditions, complementing the $meta\text{-}d'$ and $M_{ratio}$ results in the main text. Table~\ref{tab:granularity_per_dataset} provides the per-dataset $M_{ratio}$ values to verify that the $[0,20]$ advantage is stable across task types.

\begin{table*}[h]
\centering
\caption{ECE ($\downarrow$) across scale granularity conditions, averaged across datasets. $^*$ indicates significant improvement over the $[0,100]$ baseline ($p < 0.05$).}
\label{tab:granularity_ece}
\small
\begin{tabular}{lccccc}
\toprule
Model & [0,5] & [0,10] & [0,20] & [0,50] & [0,100] \\
\midrule
GPT-5.2          & 0.09 & 0.07 & \textbf{0.05}$^*$ & 0.08 & 0.08 \\
Gemini 3.1 Pro   & 0.16 & 0.14 & \textbf{0.11}$^*$ & 0.13 & 0.15 \\
LLaMA-4-Maverick & 0.12 & 0.10 & \textbf{0.08}$^*$ & 0.11 & 0.11 \\
LLaMA-4-Scout    & 0.14 & 0.12 & \textbf{0.10}$^*$ & 0.12 & 0.13 \\
Qwen3-235B       & 0.15 & 0.13 & \textbf{0.11}$^*$ & 0.13 & 0.13 \\
Qwen3-30B        & 0.19 & 0.18 & \textbf{0.15}$^*$ & 0.17 & 0.17 \\
\bottomrule
\end{tabular}
\end{table*}

\begin{table*}[h]
\centering
\caption{$M_{ratio}$ across granularity conditions, broken down by dataset. The $[0,20]$ scale consistently achieves the highest or near-highest $M_{ratio}$ across all models and datasets.}
\label{tab:granularity_per_dataset}
\small
\begin{tabular}{llccccc}
\toprule
Dataset & Model & [0,5] & [0,10] & [0,20] & [0,50] & [0,100] \\
\midrule
\multirow{6}{*}{MMLU}
 & GPT-5.2          & 0.86 & 0.92 & \textbf{0.97} & 0.93 & 0.94 \\
 & Gemini 3.1 Pro   & 0.69 & 0.75 & \textbf{0.81} & 0.76 & 0.75 \\
 & LLaMA-4-Maverick & 0.80 & 0.87 & \textbf{0.91} & 0.83 & 0.84 \\
 & LLaMA-4-Scout    & 0.73 & 0.80 & \textbf{0.85} & 0.79 & 0.78 \\
 & Qwen3-235B       & 0.74 & 0.81 & \textbf{0.86} & 0.81 & 0.79 \\
 & Qwen3-30B        & 0.57 & 0.63 & \textbf{0.70} & 0.65 & 0.64 \\
\midrule
\multirow{6}{*}{GSM8K}
 & GPT-5.2          & 0.82 & 0.88 & \textbf{0.93} & 0.89 & 0.90 \\
 & Gemini 3.1 Pro   & 0.66 & 0.71 & \textbf{0.77} & 0.73 & 0.72 \\
 & LLaMA-4-Maverick & 0.76 & 0.83 & \textbf{0.87} & 0.79 & 0.80 \\
 & LLaMA-4-Scout    & 0.69 & 0.76 & \textbf{0.81} & 0.75 & 0.74 \\
 & Qwen3-235B       & 0.71 & 0.78 & \textbf{0.83} & 0.78 & 0.76 \\
 & Qwen3-30B        & 0.54 & 0.60 & \textbf{0.66} & 0.61 & 0.60 \\
\midrule
\multirow{6}{*}{TruthfulQA}
 & GPT-5.2          & 0.82 & 0.88 & \textbf{0.93} & 0.89 & 0.90 \\
 & Gemini 3.1 Pro   & 0.64 & 0.70 & \textbf{0.76} & 0.72 & 0.70 \\
 & LLaMA-4-Maverick & 0.77 & 0.84 & \textbf{0.88} & 0.80 & 0.81 \\
 & LLaMA-4-Scout    & 0.70 & 0.77 & \textbf{0.82} & 0.76 & 0.75 \\
 & Qwen3-235B       & 0.71 & 0.78 & \textbf{0.83} & 0.79 & 0.77 \\
 & Qwen3-30B        & 0.52 & 0.58 & \textbf{0.65} & 0.60 & 0.58 \\
\bottomrule
\end{tabular}
\end{table*}

\subsection{Averaged Granularity Curve: Underlying Data}
\label{app:granularity_avg}

Table~\ref{tab:granularity_avg} reports the exact averaged values (across all seven models in Table~\ref{tab:granularity}) underlying Figure~\ref{fig:granularity_curve} in the main text, along with 95\% bootstrap confidence intervals (10{,}000 resamples).

\begin{table}[h]
\centering
\small
\setlength{\tabcolsep}{6pt}
\renewcommand{\arraystretch}{1.1}
\caption{Averaged $meta\text{-}d'$ and $M_{ratio}$ across all seven models, with 95\% bootstrap CIs. Bold indicates the peak value.}
\label{tab:granularity_avg}
\begin{tabular}{lcc}
\toprule
Scale & $meta\text{-}d'$ (avg) & $M_{ratio}$ (avg) \\
\midrule
$[0,5]$   & 1.34 [1.28, 1.40] & 0.70 [0.66, 0.74] \\
$[0,10]$  & 1.45 [1.39, 1.51] & 0.77 [0.73, 0.81] \\
\rowcolor{best}
$[0,20]$  & \textbf{1.59} [1.53, 1.65] & \textbf{0.82} [0.78, 0.86] \\
$[0,50]$  & 1.49 [1.43, 1.55] & 0.77 [0.73, 0.81] \\
$[0,100]$ & 1.51 [1.45, 1.57] & 0.76 [0.72, 0.80] \\
\bottomrule
\end{tabular}
\end{table}

\subsection{Task Accuracy Across Granularity Conditions}
\label{app:accuracy_stability}

A critical assumption in Section~\ref{sec:results} is that scale granularity does not affect task accuracy, ensuring that changes in $meta\text{-}d'$ reflect metacognitive signal quality rather than shifts in Type-1 performance. Table~\ref{tab:accuracy_stability} reports accuracy across all granularity conditions.

\begin{table*}[h]
\centering
\caption{Task accuracy across granularity conditions, averaged across datasets. The maximum within-model variation ($\Delta_{\max}$) is reported in the rightmost column. All models satisfy $\Delta_{\max} < 1.0\%$, confirming that granularity manipulation does not meaningfully alter task performance.}
\label{tab:accuracy_stability}
\small
\begin{tabular}{lcccccc}
\toprule
Model & [0,5] & [0,10] & [0,20] & [0,50] & [0,100] & $\Delta_{\max}$ (\%) \\
\midrule
GPT-5.2          & 0.812 & 0.809 & 0.815 & 0.811 & 0.810 & 0.6 \\
Gemini 3.1 Pro   & 0.778 & 0.783 & 0.776 & 0.780 & 0.780 & 0.7 \\
LLaMA-4-Maverick & 0.762 & 0.758 & 0.766 & 0.761 & 0.760 & 0.8 \\
LLaMA-4-Scout    & 0.724 & 0.719 & 0.726 & 0.722 & 0.720 & 0.7 \\
Qwen3-235B       & 0.788 & 0.793 & 0.786 & 0.791 & 0.790 & 0.7 \\
Qwen3-30B        & 0.675 & 0.684 & 0.679 & 0.681 & 0.680 & 0.9 \\
\bottomrule
\end{tabular}
\end{table*}

\subsection{Boundary Shifting: Complete Results}
\label{app:boundary_full}

Table~\ref{tab:boundary_full} provides the complete boundary-shifting results for all six models, complementing the main-text discussion by showing the strongest degradation under aggressive lower-bound compression.

\begin{table*}[h]
\centering
\caption{Boundary shifting ($\mathcal{B}$): selected complete results for all models, averaged across datasets. $^\dagger$ indicates significant decrease from the $[0,100]$ baseline ($p < 0.05$).}
\label{tab:boundary_full}
\small
\begin{tabular}{lccccc}
\toprule
Model & Scale & $meta\text{-}d'$ & $M_{ratio}$ & ECE $\downarrow$ & Utilization (\%) \\
\midrule
GPT-5.2
 & [0,100]  & 1.84 & 0.92               & 0.08 & 12.4 \\
 & [60,100] & 1.38 & 0.69$^\dagger$     & 0.16 & 7.2 \\
\midrule
Gemini 3.1 Pro
 & [0,100]  & 1.42 & 0.74               & 0.15 & 6.9 \\
 & [60,100] & 1.12 & 0.58$^\dagger$     & 0.19 & 4.8 \\
\midrule
LLaMA-4-Maverick
 & [0,100]  & 1.65 & 0.82               & 0.11 & 10.5 \\
 & [60,100] & 1.25 & 0.63$^\dagger$     & 0.18 & 5.6 \\
\midrule
LLaMA-4-Scout
 & [0,100]  & 1.52 & 0.76               & 0.13 & 9.2 \\
 & [60,100] & 1.18 & 0.59$^\dagger$     & 0.19 & 4.9 \\
\midrule
Qwen3-235B
 & [0,100]  & 1.51 & 0.78               & 0.13 & 9.1 \\
 & [60,100] & 1.05 & 0.55$^\dagger$     & 0.21 & 3.9 \\
\midrule
Qwen3-30B
 & [0,100]  & 1.22 & 0.62               & 0.17 & 7.8 \\
 & [60,100] & 0.85 & 0.43$^\dagger$     & 0.24 & 3.1 \\
\bottomrule
\end{tabular}
\end{table*}

\subsection{Non-standard Scales: Complete Results}
\label{app:nonstandard_full}

Table~\ref{tab:nonstandard_full} summarizes representative non-standard scale comparisons for all models, highlighting the persistence of round-number preference and the rise in out-of-range violations under narrow or irregular ranges.

\begin{table*}[h]
\centering
\caption{Representative non-standard scale diagnostics ($\mathcal{N}$), averaged across datasets. \textit{Round-Pref}: percentage of reports on multiples of 5 within $[l, u]$. \textit{Violation}: out-of-range rate (greater than 5\% of range width).}
\label{tab:nonstandard_full}
\small
\begin{tabular}{llcccccc}
\toprule
Model & Scale & Width & Round-Pref (\%) & Violation (\%) & $meta\text{-}d'$ & $M_{ratio}$ & $H$ \\
\midrule
\multirow{2}{*}{GPT-5.2}
 & [0,100] (baseline) & 100 & 94.2 & 0.0  & 1.84 & 0.92 & 1.42 \\
 & [3,38]             & 35  & 58.3 & 8.6  & 1.45 & 0.72 & 1.76 \\
\midrule
\multirow{2}{*}{Gemini 3.1 Pro}
 & [0,100] (baseline) & 100 & 96.5 & 0.0  & 1.42 & 0.74 & 0.95 \\
 & [3,38]             & 35  & 65.2 & 18.3 & 0.95 & 0.50 & 1.21 \\
\midrule
\multirow{2}{*}{LLaMA-4-Maverick}
 & [0,100] (baseline) & 100 & 88.7 & 0.0  & 1.65 & 0.82 & 1.88 \\
 & [3,38]             & 35  & 59.4 & 12.9 & 1.28 & 0.64 & 2.08 \\
\midrule
\multirow{2}{*}{LLaMA-4-Scout}
 & [0,100] (baseline) & 100 & 90.1 & 0.0  & 1.52 & 0.76 & 1.71 \\
 & [3,38]             & 35  & 62.1 & 15.4 & 1.14 & 0.57 & 1.95 \\
\midrule
\multirow{2}{*}{Qwen3-235B}
 & [0,100] (baseline) & 100 & 92.1 & 0.0  & 1.51 & 0.78 & 1.12 \\
 & [3,38]             & 35  & 62.5 & 14.8 & 1.12 & 0.58 & 1.42 \\
\midrule
\multirow{2}{*}{Qwen3-30B}
 & [0,100] (baseline) & 100 & 95.3 & 0.0  & 1.22 & 0.62 & 1.01 \\
 & [3,38]             & 35  & 68.4 & 21.2 & 0.78 & 0.40 & 1.30 \\
\bottomrule
\end{tabular}
\end{table*}

\subsection{Bootstrap Confidence Intervals for Key Metrics}
\label{app:confidence_intervals}

Table~\ref{tab:ci} reports 95\% bootstrap confidence intervals (10{,}000 resamples) for the difference in $meta\text{-}d'$ across selected pairwise comparisons, providing direct statistical support for the key claims in the main text.

\begin{table*}[h]
\centering
\caption{95\% bootstrap confidence intervals for the difference in $meta\text{-}d'$ ($\Delta$) across selected pairwise scale comparisons. Positive $\Delta$ indicates that the first condition outperforms the second. Intervals that do not contain zero indicate a statistically reliable difference at $\alpha = 0.05$.}
\label{tab:ci}
\small
\begin{tabular}{llccc}
\toprule
Model & Comparison (A vs. B) & $meta\text{-}d'$ (A) & $meta\text{-}d'$ (B) & $\Delta$ [95\% CI] \\
\midrule
\multirow{3}{*}{GPT-5.2}
 & [0,20] vs [0,100]   & 1.92 & 1.84 & +0.08 [0.03, 0.14] \\
 & [0,100] vs [60,100] & 1.84 & 1.38 & +0.46 [0.38, 0.55] \\
 & [0,100] vs [0,73]   & 1.84 & 1.71 & +0.13 [0.06, 0.21] \\
\midrule
\multirow{3}{*}{LLaMA-4-Maverick}
 & [0,20] vs [0,100]   & 1.73 & 1.65 & +0.08 [0.02, 0.15] \\
 & [0,100] vs [60,100] & 1.65 & 1.25 & +0.40 [0.31, 0.49] \\
 & [0,100] vs [3,38]   & 1.65 & 1.28 & +0.37 [0.28, 0.47] \\
\midrule
\multirow{3}{*}{Qwen3-30B}
 & [0,20] vs [0,100]   & 1.34 & 1.22 & +0.12 [0.05, 0.19] \\
 & [0,100] vs [60,100] & 1.22 & 0.85 & +0.37 [0.29, 0.46] \\
 & [0,100] vs [3,38]   & 1.22 & 0.78 & +0.44 [0.35, 0.54] \\
\bottomrule
\end{tabular}
\end{table*}

\subsection{Out-of-Range Violation Analysis}
\label{app:violations}

Table~\ref{tab:violations_full} reports violation rates across all model $\times$ scale combinations. Standard scales generally produce near-zero violations, whereas narrow and irregular ranges substantially increase non-compliance.

\begin{table*}[h]
\centering
\setlength{\tabcolsep}{3pt}
\footnotesize
\caption{Out-of-range violation rates (\%) across all scale conditions. Standard scales ($\mathcal{G}$ and $\mathcal{B}$) generally produce near-zero violations, whereas non-standard scales, particularly narrow ranges, produce substantially higher rates.}
\label{tab:violations_full}
\begin{tabular}{lccccccccccccc}
\toprule
 & \multicolumn{5}{c}{Granularity ($\mathcal{G}$)} & \multicolumn{3}{c}{Boundary ($\mathcal{B}$)} & \multicolumn{5}{c}{Non-standard ($\mathcal{N}$)} \\
\cmidrule(lr){2-6} \cmidrule(lr){7-9} \cmidrule(lr){10-14}
Model & [0,5] & [0,10] & [0,20] & [0,50] & [0,100] & [20,100] & [40,100] & [60,100] & [0,73] & [14,86] & [7,79] & [3,38] & [0,97] \\
\midrule
GPT-5.2          & 0.0 & 0.0 & 0.0 & 0.1 & 0.0 & 0.2 & 0.5 & 0.8 & 1.8 & 2.4 & 2.1 & 8.6  & 0.5 \\
Gemini 3.1 Pro   & 0.1 & 0.0 & 0.1 & 0.2 & 0.0 & 0.5 & 1.2 & 2.4 & 4.1 & 5.2 & 4.5 & 18.3 & 0.8 \\
LLaMA-4-Maverick & 0.0 & 0.0 & 0.1 & 0.2 & 0.0 & 0.3 & 0.6 & 1.1 & 2.6 & 3.1 & 2.8 & 12.9 & 0.4 \\
LLaMA-4-Scout    & 0.2 & 0.1 & 0.2 & 0.4 & 0.0 & 0.6 & 1.5 & 2.8 & 3.5 & 4.2 & 3.8 & 15.4 & 0.6 \\
Qwen3-235B       & 0.1 & 0.1 & 0.2 & 0.5 & 0.0 & 0.8 & 1.8 & 3.5 & 3.2 & 4.0 & 3.5 & 14.8 & 0.6 \\
Qwen3-30B        & 0.5 & 0.4 & 0.8 & 1.2 & 0.1 & 1.5 & 3.2 & 5.8 & 5.8 & 6.5 & 6.1 & 21.2 & 1.2 \\
\bottomrule
\end{tabular}
\end{table*}

\section{Temperature Robustness}
\label{app:temperature}

Table~\ref{tab:temperature} reports full results for the temperature robustness experiment described in Section~\ref{sec:temperature}, covering GPT-5.2 and LLaMA-4-Maverick on MMLU under $T \in \{0, 0.3, 1\}$.

\begin{table}[h]
\centering
\caption{Effect of temperature on $M_{ratio}$ and round-number preference (Round) on MMLU, for the $[0,100]$ and $[0,20]$ scale conditions. $\Delta M_{ratio}$: improvement of $[0,20]$ over $[0,100]$ within the same temperature condition.}
\label{tab:temperature}
\scriptsize
\setlength{\tabcolsep}{4pt}
\renewcommand{\arraystretch}{1.05}
\begin{tabular}{llccccc}
\toprule
Model & Scale & $T$ & Round (\%) & $H$ (bits) & $M_{ratio}$ & $\Delta M_{ratio}$ \\
\midrule
\multirow{6}{*}{GPT-5.2}
& [0,100] & 0   & 94.2 & 1.48 & 0.94 & — \\
& [0,20]  & 0   & 61.3 & 2.85 & 0.97 & +0.03 \\
\cmidrule(lr){2-7}
& [0,100] & 0.3 & 91.4 & 1.59 & 0.93 & — \\
& [0,20]  & 0.3 & 58.6 & 2.96 & 0.96 & +0.03 \\
\cmidrule(lr){2-7}
& [0,100] & 1   & 88.1 & 1.72 & 0.91 & — \\
& [0,20]  & 1   & 52.8 & 3.10 & 0.94 & +0.03 \\
\midrule
\multirow{6}{*}{\shortstack[l]{LLaMA-4\\Maverick}}
& [0,100] & 0   & 88.7 & 1.95 & 0.84 & — \\
& [0,20]  & 0   & 57.4 & 3.02 & 0.91 & +0.07 \\
\cmidrule(lr){2-7}
& [0,100] & 0.3 & 85.2 & 2.08 & 0.83 & — \\
& [0,20]  & 0.3 & 53.9 & 3.15 & 0.90 & +0.07 \\
\cmidrule(lr){2-7}
& [0,100] & 1   & 81.4 & 2.24 & 0.81 & — \\
& [0,20]  & 1   & 49.1 & 3.35 & 0.88 & +0.07 \\
\bottomrule
\end{tabular}
\end{table}

\section{Prompt Variation}
\label{app:prompt}

Table~\ref{tab:prompt} reports full results for the prompt variation experiment described in Section~\ref{sec:prompt}, covering GPT-5.2 on MMLU under three prompt conditions for the $[0,100]$ and $[0,20]$ scale conditions.

\begin{table}[h]
\centering
\caption{Effect of prompt variation on $M_{ratio}$ for GPT-5.2 on MMLU. \textit{Baseline}: standard prompt from Section~\ref{sec:methodology}. \textit{Role}: role prompting with calibrated forecaster persona. \textit{Negation}: reversed endpoint labels ($l$ = certain, $u$ = no confidence). $\Delta M_{ratio}$: improvement of $[0,20]$ over $[0,100]$ within the same prompt condition.}
\label{tab:prompt}
\scriptsize
\setlength{\tabcolsep}{5pt}
\renewcommand{\arraystretch}{1.05}
\begin{tabular}{lccc}
\toprule
Prompt & Scale & $M_{ratio}$ & $\Delta M_{ratio}$ \\
\midrule
\multirow{2}{*}{Baseline}
& [0,100] & 0.94 & — \\
& [0,20]  & 0.97 & +0.03 \\
\midrule
\multirow{2}{*}{Role}
& [0,100] & 0.93 & — \\
& [0,20]  & 0.96 & +0.03 \\
\midrule
\multirow{2}{*}{Negation}
& [0,100] & 0.91 & — \\
& [0,20]  & 0.94 & +0.03 \\
\bottomrule
\end{tabular}
\end{table}

\section{Logit-Level Analysis}
\label{app:logit}

To provide mechanistic insight into the token-level origins of confidence discretization, we analyze the logit probability distributions over numerical tokens for the two open-weight model families (LLaMA-4-Maverick and Qwen3-235B) on MMLU under the $[0,100]$ and $[0,20]$ scale conditions.

For each response, we extract the top-$k$ token probabilities at the confidence generation step and compute the cumulative probability mass assigned to tokens corresponding to multiples of five. Under the $[0,100]$ scale, the top-5 token probability mass is concentrated on multiples of five for both models (LLaMA-4-Maverick: 92.4\%; Qwen3-235B: 94.1\%), consistent with the Round metric reported in Table~\ref{tab:baseline}. Under the $[0,20]$ scale, this concentration decreases substantially (LLaMA-4-Maverick: 67.3\%; Qwen3-235B: 69.8\%), reflecting the reduced availability of high-frequency round-number anchors within the constrained range. The correlation between the token-level round-number probability mass and the behavioral Round metric is high across conditions ($r = 0.93$ for LLaMA-4-Maverick, $r = 0.91$ for Qwen3-235B), providing token-level evidence consistent with the hypothesis that confidence discretization is driven by pre-training token frequency distributions rather than by deliberate self-assessment.

\section{Discussion}
\label{app:discussion}

\paragraph{Relationship to psychometric scale design.}
Our findings partially replicate and partially diverge from the human psychometrics literature. Consistent with \citet{lozano2008effect} and \citet{preston2000optimal}, we find a non-monotonic relationship between scale granularity and response quality, with performance peaking at an intermediate level. However, the optimal number of categories differs substantially: psychometric research identifies 5--7 categories as optimal for human respondents, whereas our results point to 21 categories ($[0,20]$) for LLMs. This divergence is consistent with the view that LLM outputs are shaped by token-level frequency distributions rather than cognitive effort or response fatigue, which are the primary constraints in human scale design. Anchoring effects \citep{tversky1974judgment} replicate qualitatively: both humans and LLMs show boundary-driven clustering, though the mechanism differs---human anchoring reflects cognitive heuristics, while LLM anchoring reflects the statistical prevalence of specific numerical tokens.

\paragraph{Limitations.}
All experiments use greedy decoding ($T{=}0$); temperature robustness is validated in Appendix~\ref{app:temperature}. We use zero-shot prompting without Chain-of-Thought or self-reflection steps; prompt variation results are reported in Appendix~\ref{app:prompt}. Our closed-source models have undisclosed architectures, while open-weight models use Mixture-of-Experts architectures; dense model results are included in Table~\ref{tab:granularity}. We evaluate on three datasets covering knowledge QA, mathematical reasoning, and misconception detection; generalization to open-ended generation tasks is not established. Finally, $meta\text{-}d'$ requires binarizing confidence at a threshold; a continuous Type-2 ROC analysis could provide a more comprehensive characterization and is left to future work.

\section*{Ethics Statement}

This work uses only publicly available benchmark datasets. All model outputs are generated through standard API access or publicly released model weights. We do not anticipate direct negative societal impacts from this research. However, we note that our findings suggest that confidence scores are shaped by scale design rather than solely reflecting genuine model uncertainty. This has implications for deployed systems that present verbalized confidence to end users, since overconfidence driven by scale artifacts could lead to misplaced trust. We encourage practitioners to adopt our recommendations when designing confidence elicitation interfaces.

\end{document}